\pdfoutput = 1
\documentclass[letterpaper, 10pt, conference]{ieeeconf}

\IEEEoverridecommandlockouts  
\makeatletter
\let\NAT@parse\undefined
\makeatother

\usepackage{cite} 
\usepackage[pdftex]{graphics} 
\usepackage{epsfig} 
\usepackage[cmex10]{amsmath} 
\usepackage{newtxmath} 
\usepackage{times} 
\usepackage[hidelinks]{hyperref} 
\usepackage[all]{hypcap} 
\usepackage[capitalise,nameinlink]{cleveref} 
\usepackage{algorithm}
\usepackage[noend]{algpseudocode}
\usepackage{subfigure}
\usepackage{bm}
\usepackage{multirow}

\usepackage{booktabs}

\newcommand\norm[1]{\left\lVert#1\right\rVert}

\title{\LARGE \bf Comparison Study of Nonlinear Optimization of Step Durations and Foot Placement for Dynamic Walking}

\author{Wenbin Hu, Iordanis Chatzinikolaidis, Kai Yuan, and Zhibin Li
\thanks{Wenbin Hu is an undergraduate student with the Department of Automation, Tsinghua University, China. Other authors are with the University of Edinburgh.
{Email: hwb14@mails.tsinghua.edu.cn}}
}

\begin{document}

\maketitle
\thispagestyle{empty}
\pagestyle{empty}

\begin{abstract}
This paper studies bipedal locomotion as a nonlinear optimization problem based on continuous and discrete dynamics, by simultaneously optimizing the remaining step duration, the next step duration and the foot location to achieve robustness. The linear inverted pendulum as the motion model captures the center of mass dynamics and its low-dimensionality makes the problem more tractable. We first formulate a holistic approach to search for optimality in the three-dimensional parametric space and use these results as baseline. To further improve computational efficiency, our study investigates a sequential approach with two stages of customized optimization that first optimizes the current step duration, and subsequently the duration and location of the next step. The effectiveness of both approaches is successfully demonstrated in simulation by applying different perturbations. The comparison study shows that these two approaches find mostly the same optimal solutions, but the latter requires considerably less computational time, which suggests that the proposed sequential approach is well suited for real-time implementation with a minor trade-off in optimality.
\end{abstract}

\section{Introduction}

All-terrain mobility can be achieved by legged machines but maintaining their balance is challenging, especially for bipedal robots. Though minor disturbances can be handled by regulating body posture and ankle torque \cite{Stephens2007}, to ensure robust locomotion in the presence of large perturbations, the capability of changing both step duration and foot placement is very essential. This raises the demand of fast and efficient computations that permit rapid responses.

Recent works have proposed some solutions for adjusting foot placement. For this purpose, various Model Predictive Control (MPC) schemes have been proposed: control of the step positions and rotations by utilizing the time-varying Divergent Component of Motion (DCM) was conducted in \cite{MPC_0}; step position adaptation in an MPC scheme was studied in \cite{MPC_1}, \cite{MPC_2}; footstep planning optimizing the next foot placement to minimize the CoM tracking errors, was proposed in \cite{Feng}; a reactive motion controller which modifies the foot placement using MPC in the presence of a large disturbance was introduced in \cite{castano2016dynamic}. Model discrepancies were minimized in \cite{li2017robust} using optimization for accurate foot placement control.
All these methods only adjust the upcoming step location and assume a constant step duration, and therefore limit the performance since the step duration has a strong impact on the robustness of the gait recovery.

Hence, other works have studied the optimization of step duration. However, a variable step duration renders the problem nonlinear. An MPC scheme that combined ankle, hip, and stepping strategies with modifications of step time and location was proposed in \cite{strategies}. Furthermore, nonlinear optimization of multiple-step times and locations was conducted in \cite{zhou2017overview} for gait planning, and a similar scheme for the step location and timing of the present and next two steps was presented in \cite{nonlinear} for online control. 
A quadratic programming approach for real-time step location and timing adaptation based on the Linear Inverted Pendulum (LIP) model with divergent component of motion was introduced in \cite{DCM_quadratic}. The work in \cite{you2016foot} used iterative search with forward simulation of a nonlinear multi-link model to achieve dynamic walking that is more natural and human-like than that of the LIP model. 

The realization of robust walking is indeed hard because of its complex nature of continuous and discrete dynamics, i.e. the dynamics within a step (current step duration) and in-between steps (next step duration and location). To restore a stable gait, walking parameters, i.e. the remaining duration of the current step and the duration and location of the next step, need to be optimized.
The variable duration for both the current and the next step makes the problem nonlinear, because the state space representation of the motion model is changing. 
Our goal is to minimize the error of the final state of each step, i.e. both the position and velocity of the center of mass (CoM), rather than the norm of errors of the time based CoM trajectory. 

Although legged mobility has greater potential over wheeled solutions for universal traversability and has been studied for decades, there is no imminent application yet, because of the lack of locomotion robustness, high power demand from computation, etc. Hence, our study is motivated in first developing an effective problem formulation that can be solved by optimization, and then further customize the optimization procedure by the underlying physics principle, such that it becomes more computationally efficient and viable for real-world implementations.


Therefore, we investigate two proposed approaches in a sagittal scenario, and compare the disturbance rejection performance and computational time. The first approach is the canonical formulation that simultaneously optimizes the three walking parameters using a nonlinear solver. In the second approach, we decouple the original problem into two stages.
In the first stage, we optimize the current remaining step duration according to the predicted final state of the current step. In the second stage, the duration and location of the next step are optimized.
Our study shows that these two approaches exhibit the same robustness and convergence in our case studies, and only differ occasionally in the parameter scan of the CoM state.
The minor differences between the two are compared and presented in \cref{seq:simulations}.
As a result, all simulations demonstrate that both proposed solutions enable robust walking against large external disturbances.

This paper consists of the following parts.
In \cref{seq:background} the dynamics of the motion model is briefly revisited.
\cref{seq:nonlin_opt} explains the algorithmic details of the problem formulation of two optimization approaches, and introduces the notion of critical CoM states. \cref{seq:simulations} presents simulation results demonstrating the effectiveness and robustness of both approaches, and compares the subtle variation in the optimal solution, i.e. preference of long/short duration steps, and the drastic difference in computational time.
In \cref{seq:conclusion} we conclude this study and suggest future extensions.

\section{Preliminaries} \label{seq:background}


The LIP model is one of the most common templates that has been extensively used in dynamic walking control. It models the biped robot as a single point mass with a massless telescopic leg based on three assumptions: 1. the height of the center of mass remains constant; 2. the whole-body dynamics is approximated by the point mass dynamics of the CoM; 3. the contact area between the robot and the surface is a point at the foot placement with zero torque.
With these assumptions, the dynamic motion represented by the LIP model is linearized and equivalent in both the sagittal and the lateral plane. This paper studies the nonlinear optimization of foot placement control in the sagittal plane before the extension to the 3D case.

\begin{figure}[t]
  \centering
  \includegraphics[width = 0.45\textwidth]{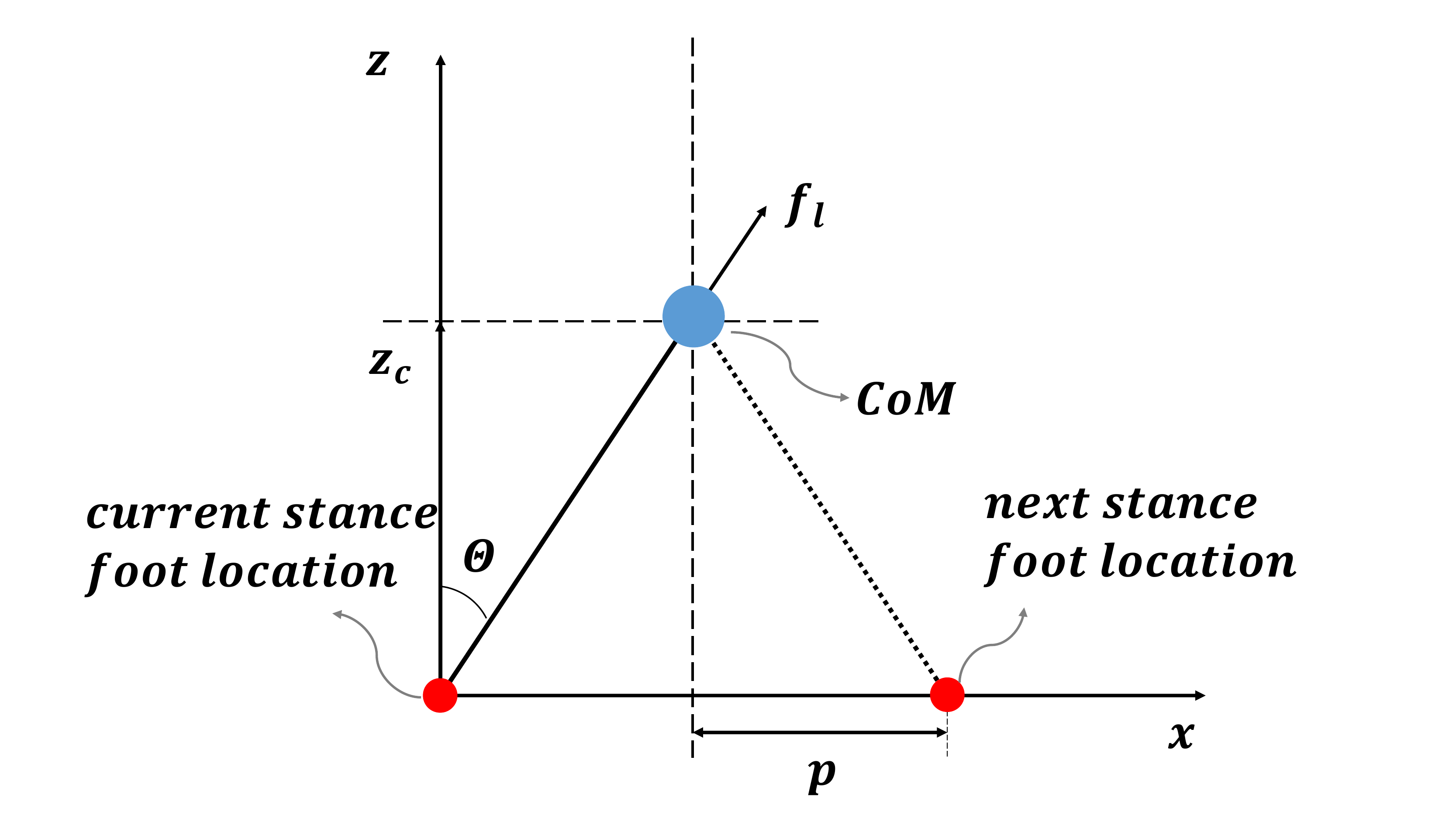}
  \caption{Dynamics of the Linear Inverted Pendulum (LIP) model.}
  \label{LIP}
\end{figure}

\cref{LIP} shows the LIP model in the local reference frame of the stance foot. The relative position of the stance foot with respect to the CoM is denoted as $p$, pointing from the horizontal projection of the CoM to the stance foot.
The axial force $f_l$ along the leg points from the foot placement location to the CoM. $\theta$ is the angle from the vertical.
To keep the height of the CoM constant at $z_c$, the vertical component of axial force must counterbalance the gravity at the CoM $f_g$, i.e. $f_l\cos{\theta} = f_g$.
The horizontal component of the axial force is
$M\ddot{x} = f_l\sin{\theta}$.
Thus, the CoM dynamics in the sagittal plane are
\begin{equation} \label{CoM dynamics}
  \ddot{x} = \frac{g}{z_c}x,
\end{equation}
where $x$ and $z_c$ are the horizontal and vertical position of the CoM in the sagittal plane, respectively.

By solving \cref{CoM dynamics}, we obtain
\begin{align}
        x(t) & = x(0)\cosh{\left(\tfrac{t}{T_c}\right)} + T_c\dot{x}(0)\sinh{\left(\tfrac{t}{T_c}\right)} \label{eqn:CoM position}\\
  \dot{x}(t) & = \tfrac{x(0)}{T_c}\sinh{\left(\tfrac{t}{T_c}\right)} + \dot{x}(0)\cosh{\left(\tfrac{t}{T_c}\right)}, \label{eqn:com velocity}
\end{align}
where $T_c = \sqrt{z_c/g}$ is the time constant, determined by the CoM height and the gravity acceleration.
\cref{eqn:CoM position,eqn:com velocity} show how the CoM state evolves after time $t$, from the initial state $\begin{bmatrix} x(0) & \dot{x}(0) \end{bmatrix}^T$ to the final state $\begin{bmatrix} x(t) & \dot{x}(t) \end{bmatrix}^T$; we use vector $\bm{x}$ to denote the CoM state $\begin{bmatrix} x & \dot{x} \end{bmatrix}^T$.

%

\section{Nonlinear optimization} \label{seq:nonlin_opt}

\subsection{Holistic approach} 

Given the current CoM position and velocity, we optimize the remaining duration of the current step $T_{s0}$, the next foot placement $p$, and the next step duration $T_{s1}$ simultaneously.

During stable walking, the LIP motion has gait symmetry about the vertical axis in \cref{LIP} within a step, i.e. $x(0) = -x_d$ and $\dot{x}(0) = \dot{x}_d$ in \cref{eqn:CoM position,eqn:com velocity}, where the subscript $d$ represents the desired CoM state at the end of a step.

For simplicity, we define $C_{i} = \cosh(T_{s_i}/T_c)$, $S_i = \sinh (T_{s_i}/T_C)$, where $i = \{0,1,d\}$.
Given the desired step duration $T_{s_d}$ and final CoM velocity $\dot{x}_d$ reached at the end of a step, the desired final state of a step can be computed as
\begin{equation} \label{Xd}
  \bm{x}_d = \begin{bmatrix} x_d \\ \dot{x}_d \end{bmatrix} = \begin{bmatrix} 
  T_c S_d/(1+ C_d)
  \\ 1 \end{bmatrix} \dot{x}_d.
\end{equation}

Given the current state $\bm{x}_0$, using \cref{eqn:CoM position,eqn:com velocity}, the final state of the current step $\bm{x}_1$ can be predicted as:
\begin{equation} \label{X1}
  \bm{x}_1 = \begin{bmatrix} x_1 \\ \dot{x}_1 \end{bmatrix} = \begin{bmatrix} C_0 & T_c S_0 \\ S_0/T_c & C_0 \end{bmatrix} \bm{x}_0.
\end{equation}

Particularly, since we focus on free walking in unconfined spaces, we define the relative foothold with respect to the CoM position as $p$. The evolution of the CoM state expressed in the local coordinate at $p$ is simplified as:
\begin{equation} \label{X2}
  \bm{x}_2 = \begin{bmatrix} -C_1 \\ -S_1/T_c \end{bmatrix}p + \begin{bmatrix} T_c S_1 \\ C_1 \end{bmatrix}\dot{x}_1 .
\end{equation}

The nonlinear optimization problem can be formulated as
\begin{equation} \label{eqn:holisitic_opt}
	\begin{IEEEeqnarraybox}[][c]{l?l}
  	\underset{\bm{v}}{\text{minimize}} & \sum_{i = 1,2} \norm{\bm{x}_i(\bm{v}) - \bm{x}_d}_{W_i}^2 \\
                     \text{subject to} & \bm{v}_{\text{min}} \leq \bm{v} \leq \bm{v}_{\text{max}},
	\end{IEEEeqnarraybox}
\end{equation}
where the cost function is the weighted norm of the CoM state errors at the end of the current and next steps, $\bm{v} \in \mathbb{R}^3$ is the vector of walking parameters to be optimized,
\begin{equation}
  \bm{v} = \begin{bmatrix} T_{s0} & T_{s1} & p \end{bmatrix}^T,
\end{equation}
and $\bm{v}_{\text{min}}$ and $\bm{v}_{\text{max}}$ are the lower and upper bounds of the optimization parameters.
Due to limited acceleration, the swing foot cannot be changed too quickly.
Therefore, this constraint is represented by the minimum bound of the step duration (swing time).
The constraints on the step length are defined by the friction cone $\theta_m$ and the height of the CoM $z_c$.

Optimization of more steps is not very helpful in feedback control since the uncertainty from sensory noises increase drastically as prediction gets longer. Moreover, our proposed optimization is updated at each control loop to mitigate any disturbances.

\subsection{Sequential approach}

\begin{figure}
	\centering
	\subfigure[$p_{\text{ls}} \leq -SL_{\text{max}} $] {
		\includegraphics[scale=0.22]{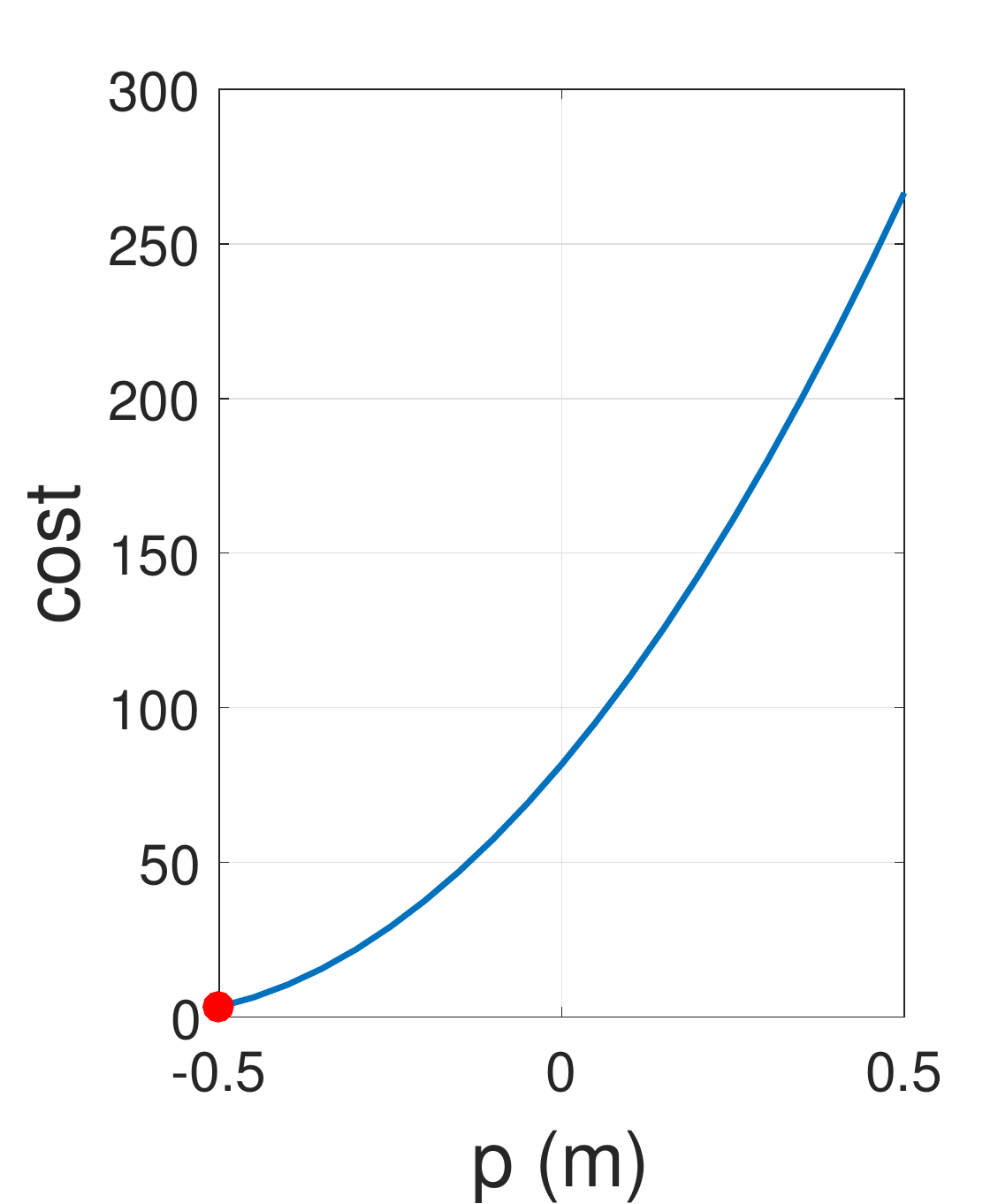}
	}
	\centering
	\subfigure[$|p_{\text{ls}}|<|L_{\text{max}}|$] {
		\includegraphics[scale=0.22]{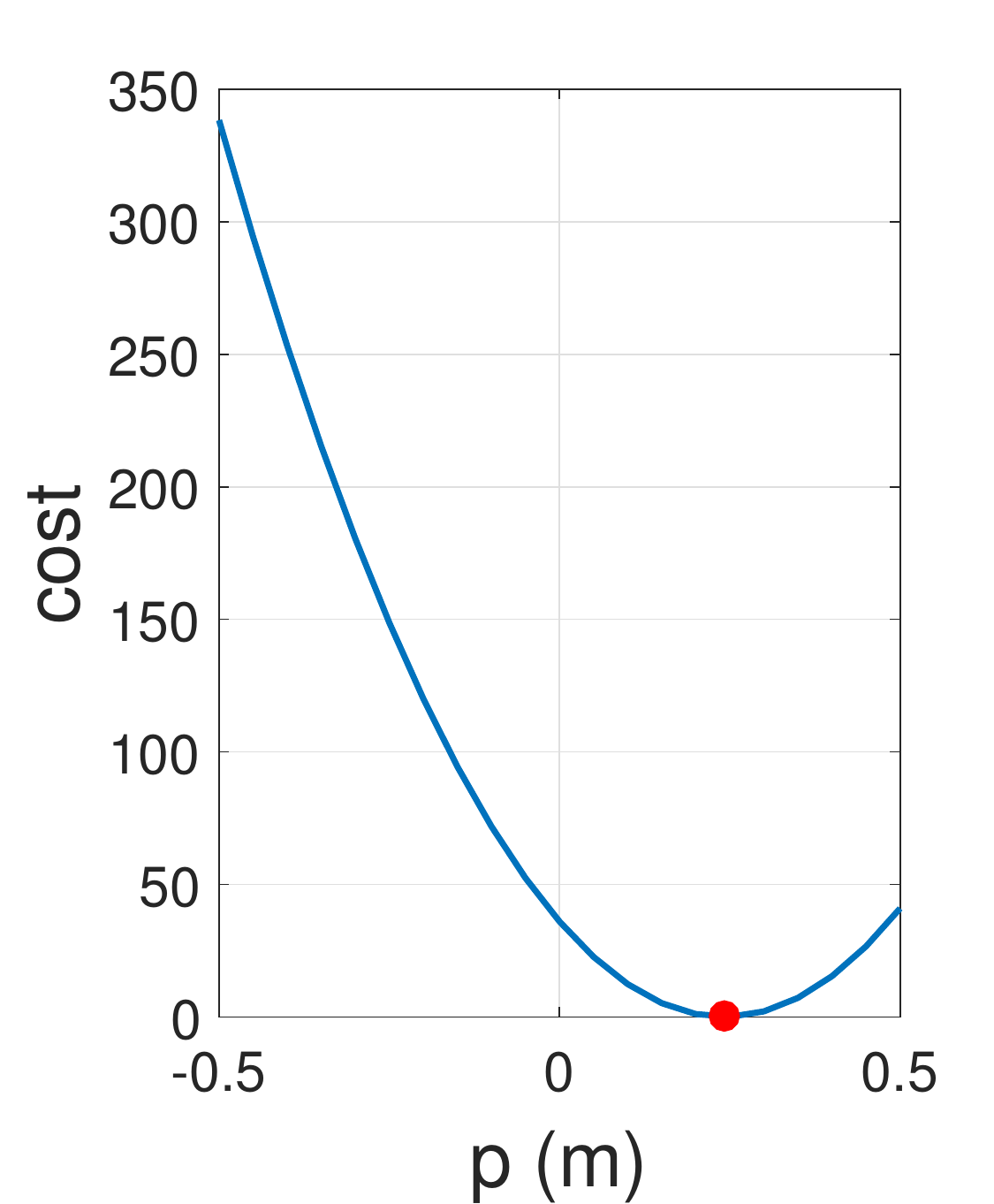}
	}
	\centering
	\subfigure[$p_{\text{ls}} \geq L_{\text{max}}$] {
		\includegraphics[scale=0.22]{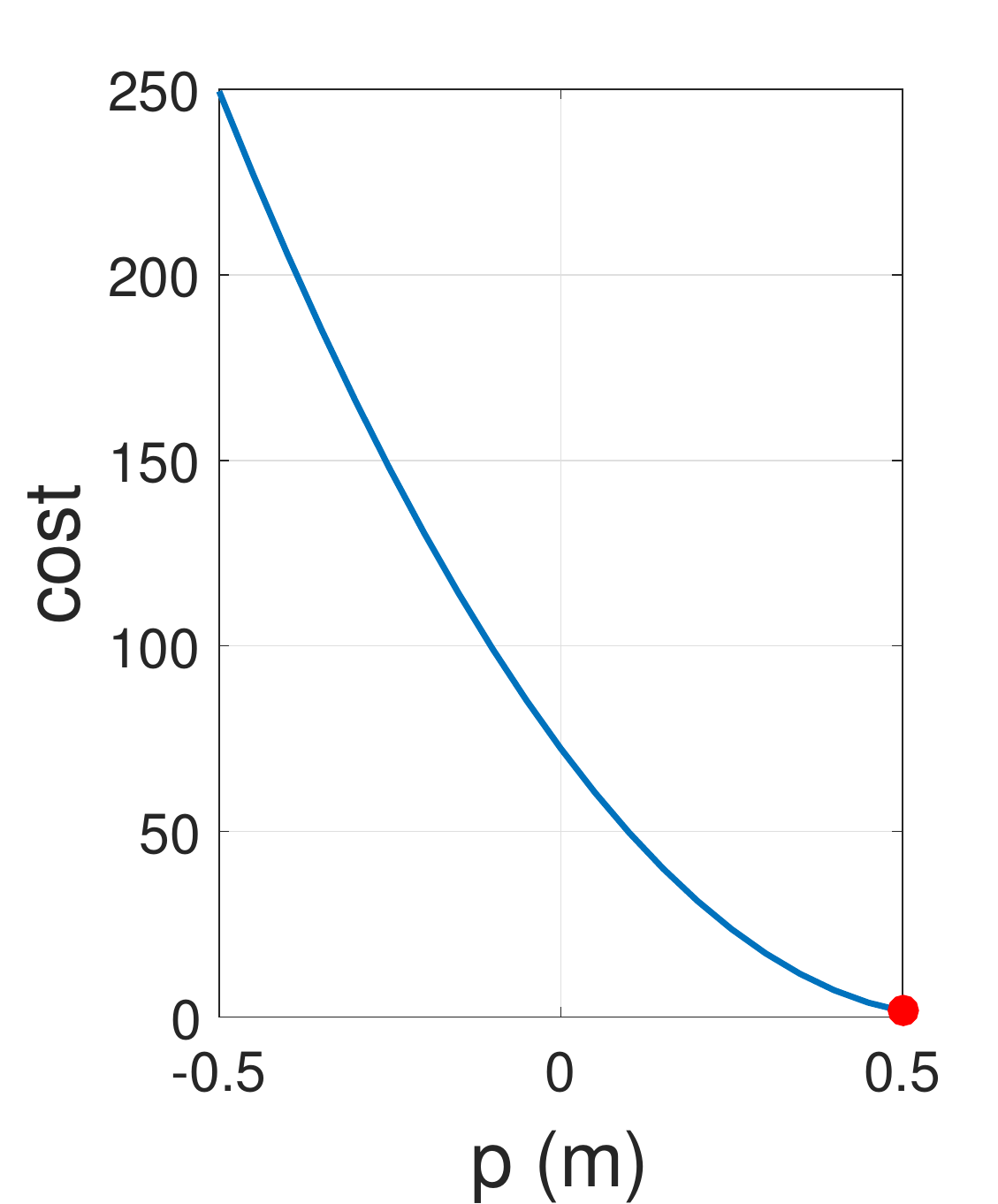}
	}
	\caption{These figures demonstrate three different situations of \cref{final p result in seq}. The costs are calculated by the objective of \cref{eqn:second_stage} when $T_{s1}$ is fixed. Here, $L_\text{max}$ is $0.5m$. The red points demonstrate $p^*$ obtained by \cref{final p result in seq} within the feasible range and the corresponding costs. }
	\label{fig:3 situations of p}
\end{figure}

Motivated by real-time applications, a customized two-stage optimization approach will be studied in the following. The idea is based on the underlying physics principles: First, there exists a remaining duration such that the final state error is minimal at the current step; second, with an initial velocity of the next step, for a given step duration within the constraint, there exists only one unique step length that results in minimal final state error while satisfying the constraint friction cone (\cref{fig:3 situations of p}). Thus, we can embed the optimization of step location with constraint inside the optimal search of the next step duration, and thus find a unique optimal set of step duration and location. In short, the nature of the model dynamics suggests that each of these optimization processes is 1-dimensional and each cost function is unimodal.

\subsubsection*{Stage 1}
The objective of the first stage is to find the optimal remaining duration that minimizes the error of the final state of the current step $\bm{x}_1$ as:
\begin{equation} \label{eqn:first_stage}
	\begin{IEEEeqnarraybox}[][c]{l?l}
	  \underset{T_{s0}}{\text{minimize}} & \norm{\bm{x}_1(T_{s0})-\bm{x}_d}_{\bm{W}}^2 \\
	                   \text{subject to} & \text{max} \{ T_{\text{min}}-T_{\text{elap}}, 0 \} \leq T_{s0} \leq T_{\text{max}},
	\end{IEEEeqnarraybox}
\end{equation}
where $T_{\text{min}}$ and $T_{\text{max}}$ are the constraints of step duration, and $T_{\text{elap}}$ is the elapsed time of the current step, and $\bm{x}_d$ and $\bm{x}_1$ are given by \cref{Xd,X1}, and $\bm{W}$ is a diagonal weighting matrix for the CoM state error of the current step.

\subsubsection*{Stage 2}

\begin{figure}[t] \label{seq_t0_middle}
  \centering
  \includegraphics[width=75mm]{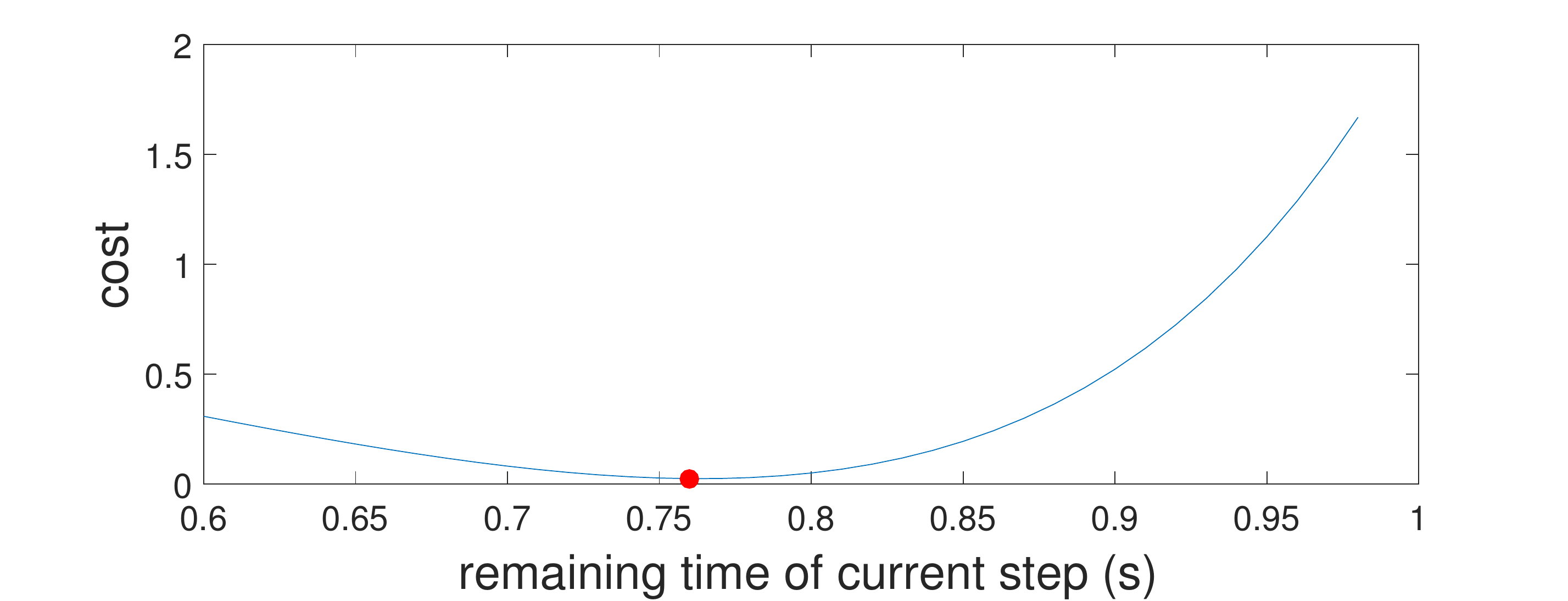}
  \caption[Stage 1 of the sequential approach]{The result of Stage 1 of the sequential approach when the CoM state is $\boldsymbol x = [0.015, -0.3]^T$.}
\end{figure}
\begin{figure}[t] \label{seq_t1_middle}
  \centering
  \includegraphics[width=8cm,height=3.5cm]{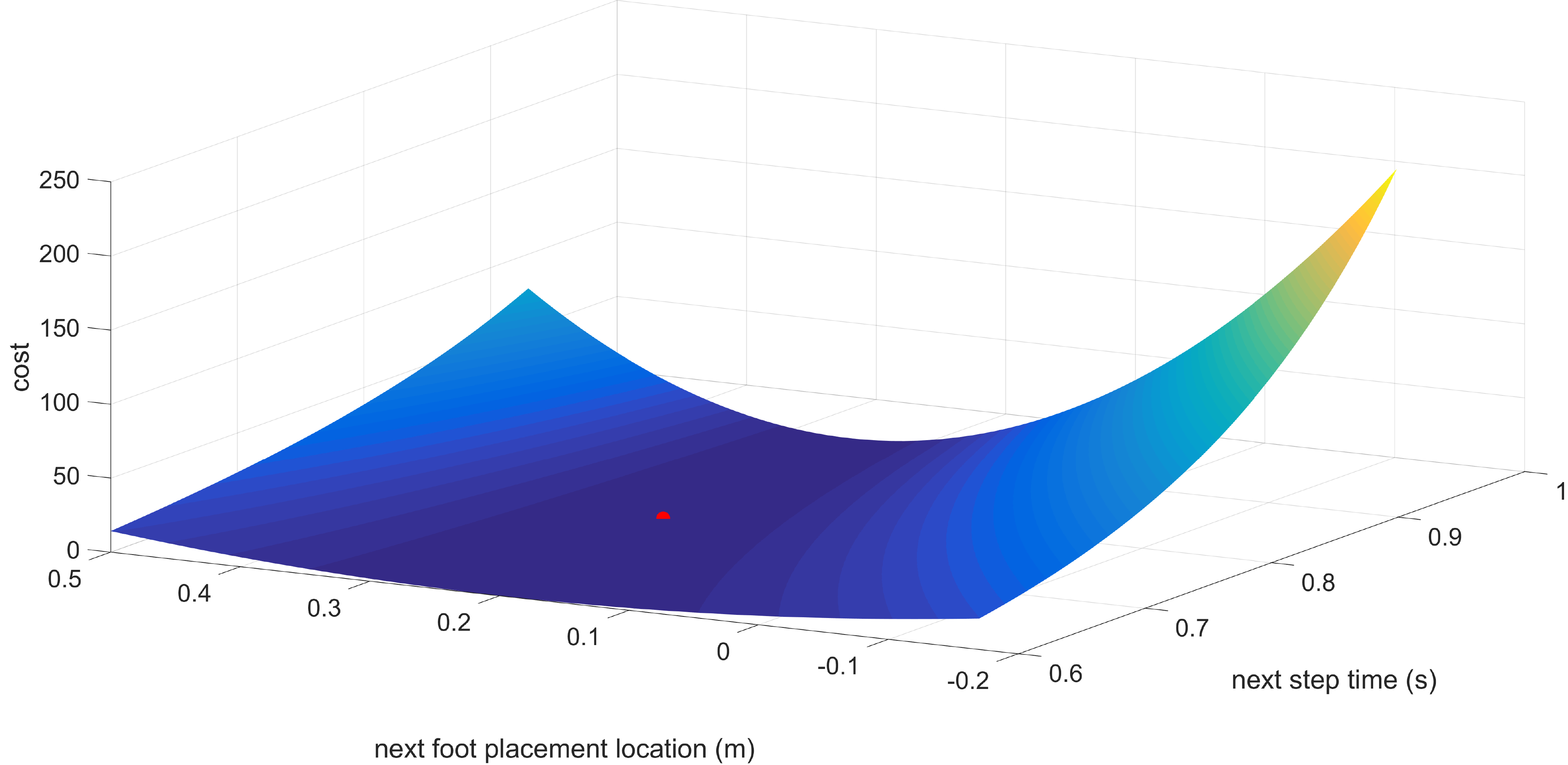}
  \caption[Stage 1 of the sequential approach]{The result of Stage 2 of the sequential approach when the CoM state is $\boldsymbol x = [0.015, -0.3]^T$. The cost function surface is based on the optimal remaining duration of the current step solved in Stage 1.}
\end{figure}

After obtaining the optimal remaining duration of the current step, we can estimate the final state of current step using \cref{X1}, where the final velocity is also the initial velocity of the next step.
Then we optimize the step duration and foot placement location of the next step.

In this stage, the optimization problem originally contains two parameters, namely the foot placement location $p$ and the next step duration $T_{s1}$. To speed up the computation, instead of optimization two parameters by canonical constrained optimization, we analytically express $p$ given the next step duration $T_{s1}$ using the weighted least squares solution first.
By adding $\bm{x}_d$ to both sides of \cref{X2} we obtain that
\begin{equation} \label{tricky seq}
  \bm{x}_d = \begin{bmatrix} -C_1 \\ -S_1/T_c \end{bmatrix} p + \begin{bmatrix} T_c S_1 \\ C_1 \end{bmatrix} \dot{x}_1 + (\bm{x}_d - \bm{x}_2).
\end{equation}
We define
\begin{IEEEeqnarray}{lCr}
\bm{A} = \begin{bmatrix} -C_1 \\ -S_1/T_c \end{bmatrix}, \ \ &
\bm{b} = \bm{x}_d - \begin{bmatrix} T_c S_1 \\ C_1 \end{bmatrix} \dot{x}_1, 
\ \ &
\bm{e} = \bm{x}_d - \bm{x}_2,
\end{IEEEeqnarray}
where $\bm{e}$ is the error between the final state of next step and the desired state.
\cref{tricky seq} can now be written as
\begin{equation} \label{simplified popt}
  \bm{b} = \bm{A} p + \bm{e}.
\end{equation}

To minimize the absolute value of $\bm{e}$, we express $p_{\text{ls}}$ by $T_{s1}$ using the weighted least squares solution
\begin{equation} \label{p in sequential method}
p_{\text{ls}} = (\bm{A}^T \bm{Q} \bm{A})^{-1} \bm{A}^T \bm{Q} (\bm{b} - \bm{e}),
\end{equation}
where $\bm{Q}$ is a weighting matrix.

Given any step duration $T_{s1}$, we can always obtain a corresponding foot placement location $p_{\text{ls}}$ using \cref{p in sequential method} which minimizes the error between $\bm{x}_2$ and $\bm{x}_d$ in an unconstrained setting.
Since the step length is limited, the foot placement location must lie within the feasible range respecting the friction. Because of the particular feature of LIP model dynamics, the cost is a unimodal function, the optimal step length $p^*$ subject to the constraint can be obtained via clipping given the unconstrained solution in \eqref{p in sequential method}:
\begin{equation} \label{final p result in seq}
	p^* = \left\{ \,
	\begin{IEEEeqnarraybox}[\IEEEeqnarraystrutmode][c]{r?s}
		-L_{\text{max}}, & if $p_{\text{ls}} \leq -L_{\text{max}}$ \\
		  p_{\text{ls}}, & if $-L_{\text{max}} < p_{\text{ls}} < L_{\text{max}}$ \\
		 L_{\text{max}}, & if $p_{\text{ls}} \geq L_{\text{max}}$
	\end{IEEEeqnarraybox}
	\right.
\end{equation}
where $L_{\text{max}}$ stands for the maximum step length.

The optimization problem of this stage is expressed as
\begin{equation} \label{eqn:second_stage}
	\begin{IEEEeqnarraybox}[][c]{l?l}
		\underset{T_{s1}}{\text{minimize}}  & \norm{\bm{x}_2(T_{s1}, p^*) - \bm{x}_d}_{\bm{W}}^2 \\
	                      \text{subject to} & T_{\text{min}} \leq T_{s1} \leq T_{\text{max}},
	\end{IEEEeqnarraybox}
\end{equation}
where $\bm{x}_d$, $\bm{x}_2$ and $p^*$ are given by \cref{final p result in seq,Xd,X2}, and $\bm{W}$ is the same weighting matrix for the CoM state error at the next step. \cref{alg:sequential} summarizes the sequential approach.

In general, adding constraints on $p$ in \cref{final p result in seq} might lead to a suboptimal result, but by fixing the $T_{s1}$, the objective of \cref{eqn:second_stage} is a single minimum function of $p$, the final result of $p^*$ obtained by \cref{final p result in seq} is always optimal within the feasible range. \cref{fig:3 situations of p} demonstrate three different cases of \cref{final p result in seq}. From the figures we can tell that the final result of $p$ is always the minimum in the feasible range.

\cref{seq_t0_middle,seq_t1_middle} present an example of how the sequential approach searches for the optimal walking parameters.
In \cref{seq_t0_middle}, $y$ is the value of the objective defined in \cref{eqn:first_stage}, while in \cref{seq_t1_middle} $y$ is the value of the objective defined in \cref{eqn:second_stage}.
The red points in figures represent the optimal results.
Both stages in the sequential approach are one-parameter nonlinear optimization problems, while the holistic approach is a three-parameter nonlinear optimization problem.
Alternatively, instead of the customized optimization consists of determining foot position by \cref{p in sequential method} first and then finding a suitable step duration by \cref{eqn:second_stage}, in fact, stage 2 can directly optimize over the next step position and duration using constrained optimization. However, by doing so, the advantage of fast solvability of the sequential approach is lost.

\begin{algorithm}[t]
  \caption{Optimization by the sequential approach.}
	\label{alg:sequential}
	\begin{algorithmic}[1]
		\Require Desired CoM state at the end of a step $\bm{x}_d$, current CoM state $\bm{x}_0$.
		\Ensure Optimizing the remaining duration of current step $T_{s0}$, step duration $T_{s1}$ and location $p^*$ of the next step. \\
		\textbf{While}{ `not fall',} \textbf{do}
			\State \ \ Obtain optimal $T_{s0}$ by solving \cref{eqn:first_stage}.
			\State \ \ Predict the final state of the current step $\bm{x}_1$ given \cref{X1}.
			\State \ \ Obtain $p^*$ given $T_{s1}$ using \cref{p in sequential method,final p result in seq}.
			\State \ \ Obtain optimal search of $T_{s1}$ by solving \cref{eqn:second_stage}.\\
		\textbf{end}
	\end{algorithmic}
\end{algorithm}

\subsection{Critical center of mass states}

If the walking parameters change continuously and smoothly as the CoM state evolves, interpolation of the walking parameters using a lookup table can provide relatively accurate results.
While generally these parameters exhibit such properties, there are some discontinuous edges, where the values of the walking parameters differ a lot on two sides.
We call these \textit{critical states}.
When encountering external disturbances near the critical states, the bipedal robot has to significantly adjust the walking pattern in order to maintain balance, even if the state of the CoM changes slightly.
Disturbances near these critical states are very dangerous and should be treated separately.

There are two reasons explaining the existence of these critical states. One is related to the orbital energy. Kajita introduced the concept of orbital energy in bipedal walking \cite{orbital_energy}, which is calculated as
\begin{equation} \label{energy}
  E = \frac{1}{2}\dot{x}^2 - \frac{g}{2z_c}x^2.
\end{equation}

According to the LIP model, the orbital energy is conserved within each step, assuming no further injection of energy from external pushes, since the disturbance is stochastic and may or may not persist.
If the CoM can move over the stance foot without reversing traveling direction, then the orbital energy is positive $E = \dot{x}_0^2/2$.
If the CoM cannot pass the stance foot, i.e. the pivot, it will stop and reverse its motion direction.
Then, the orbital energy is negative $E = -gx_{\text{rev}}^2/2z_c$, where $x_{\text{rev}}$ is the location where the CoM stops and reverses motion direction.
When the orbital energy is zero, the CoM will stop exactly above the foot placement location.
In this case, we have
\begin{equation}
  \frac{1}{2}\dot{x}^2 = \frac{g}{2z_c}x^2.
\end{equation}
When $x$ and $\dot{x}$ have opposite signs, then
\begin{equation} \label{critical states line}
  x = -\sqrt{\frac{z_c}{g}}\dot{x}.
\end{equation}

\begin{figure}[t] \label{state plane}
	\centering
	\includegraphics[width = 70mm]{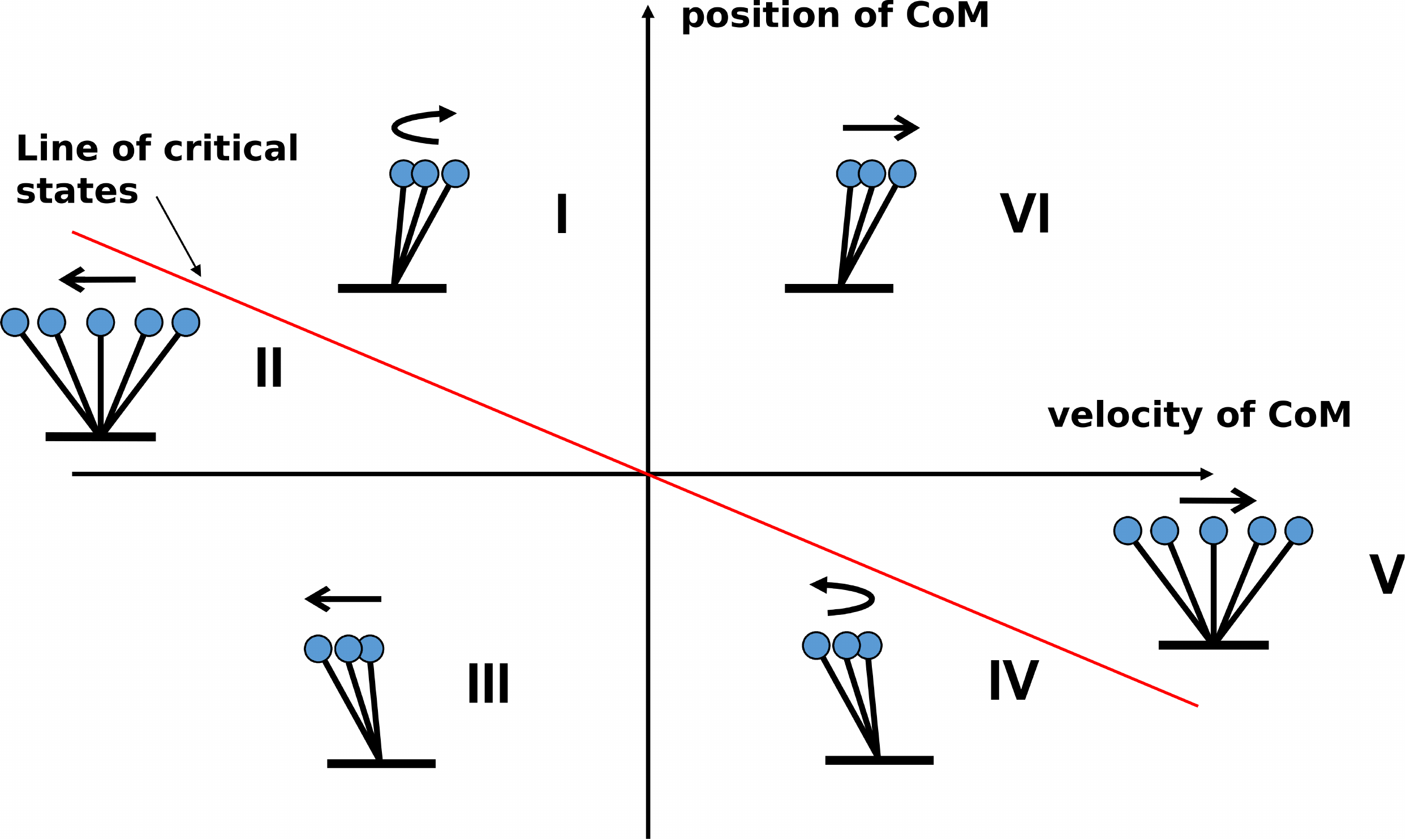}
	\caption{Convergent and divergent motions on different parts of the CoM state space.}
\end{figure}

\cref{critical states line} can be depicted as a straight line passing through the origin in the state plane, dividing the plane into several parts of interest.
Based on the signs of the CoM position, velocity, and orbital energy, we can obtain different CoM motions, as demonstrated in \cref{state plane}.
In parts II, III, and IV, the robot will eventually fall back, while in the other state space, the robot will keep moving forward.
Therefore, the states on the line of \cref{critical states line} are the critical states, or the ``\textit{twilight zone}'', because the LIP motions on the opposite side are completely different.

\begin{figure}[t] \label{top T0}
	\centering
	\includegraphics[width = 75mm]{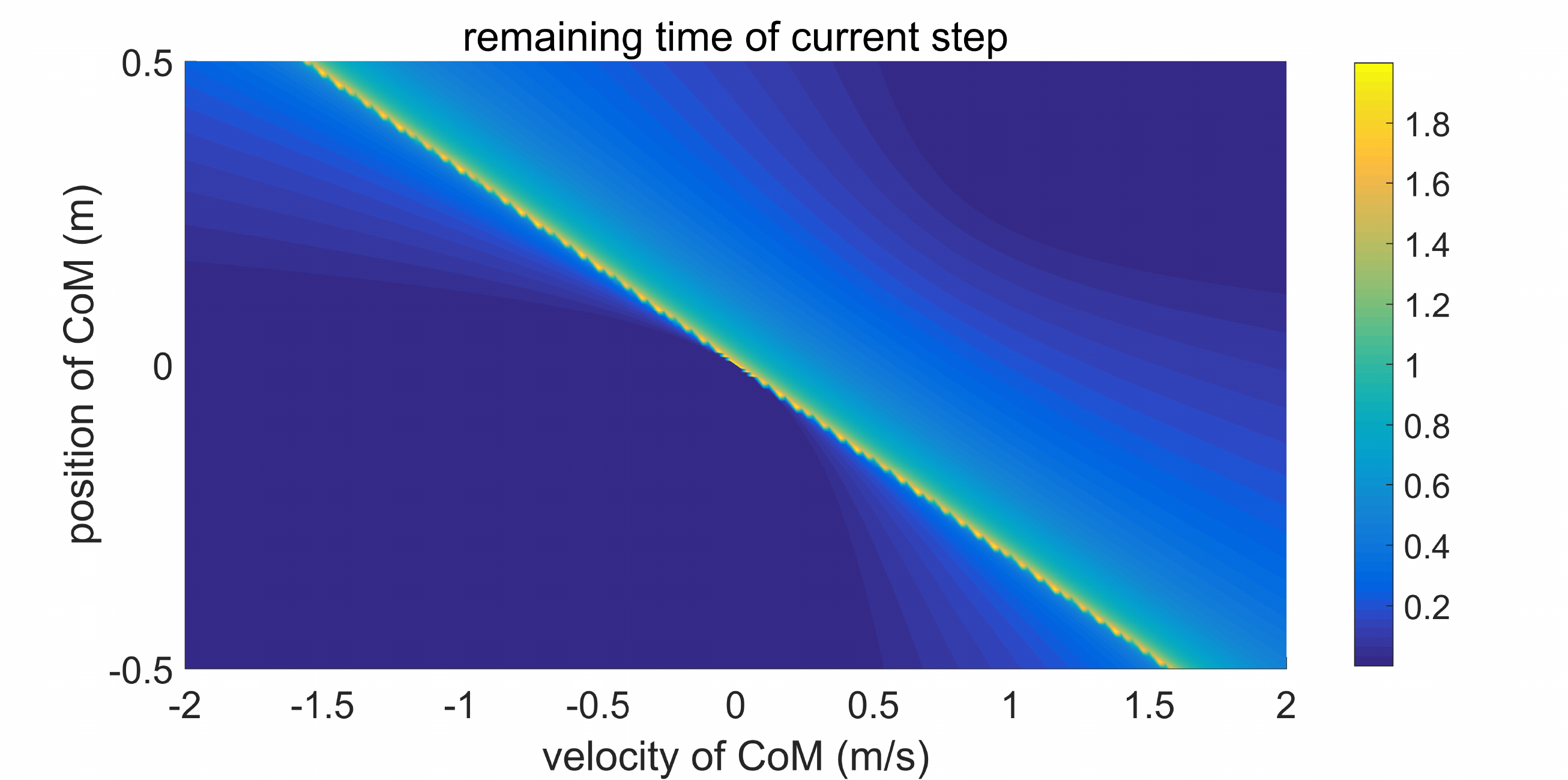}
	\caption{Top view of the 3-D figure of the optimal current remaining step duration at different discrete CoM states.}
\end{figure}

\begin{figure}[t] \label{top T1}
	\centering
	\includegraphics[width = 75mm]{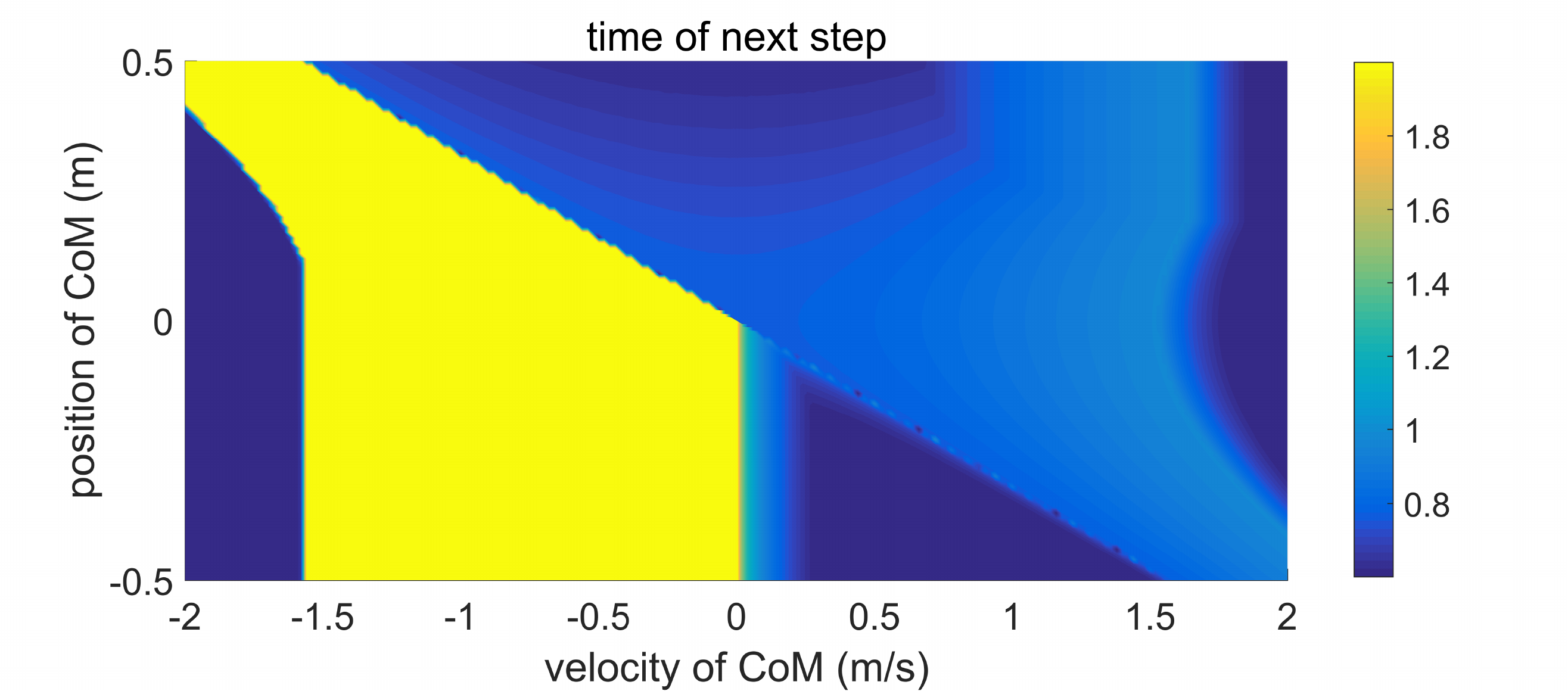}
	\caption{Top view of the 3-D figure of the optimal next step duration at different discrete CoM states.}
\end{figure}

\begin{figure}[t!] \label{top p}
	\centering
	\includegraphics[width = 75mm]{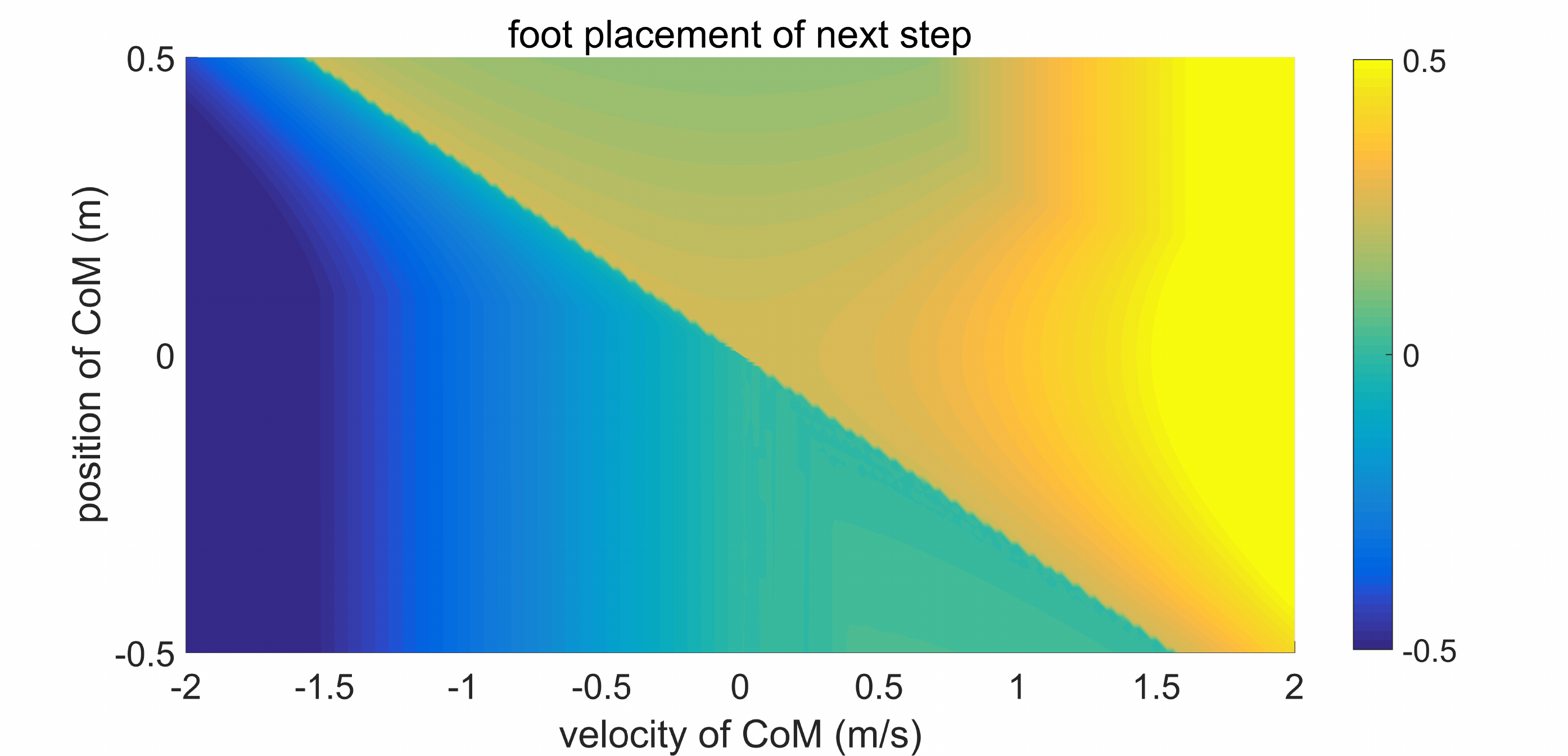}
	\caption{Top view of the 3-D figure of the optimal next foot placement at different discrete CoM states.}
\end{figure}

In \cref{top T0,top T1,top p}, the lines of the critical states can be clearly seen.
As an example, \cref{table: sample point} demonstrates the optimal walking parameters of two neighboring discrete states near this line.
There is only one centimeter difference in the CoM position between these states, but the optimal set of walking parameters is considerably different.
For $\boldsymbol x = [0.03, -0.12]^T$, the robot has a tendency of falling back, so it must take an immediate small step backwards to support itself.
For $\boldsymbol x = [0.04, -0.12]^T$, the CoM position is slightly more forward, so the robot can recover the traveling velocity by itself without stepping back, and keep moving forward steadily.

The same conclusion can be drawn from analyzing the Divergent Component of Motion \cite{DCM} which is equivalent to the Capture Point \cite{pratt_CP}. Their derivation is from the same LIP dynamics and yields the same representation as \eqref{critical states line}.


Another reason for the existence of these critical states is related to the limits of the parameters.
The sudden change of the next step duration in the left part of \cref{top T1} is because the next foot placement location reaches the lower boundary at those states, as shown in \cref{top p}.

\begin{table}[t]
  \caption{Optimal walking parameters of two neighbouring discrete states near the critical state.}
  \label{table: sample point}
  	\begin{tabular}{l |  c c}
  	&  \multicolumn{2}{c}{CoM state $\boldsymbol x$}\\ 
     &$[0.03, -0.12]^T$& $[0.04, -0.12]^T$\\
    \hline
    Current remaining step duration &   0.309 & 1.891 \\
    Next foot placement location   & -0.025 & 0.266 \\
    Next step duration&2.000&0.795\\
  \end{tabular}
\end{table}

\section{Simulation results} \label{seq:simulations}

To illustrate the effectiveness and robustness of these two nonlinear optimization approaches, we perform several simulation in MATLAB.
According to Newton's second law, external forces acting on the CoM will result in the sudden change of acceleration.
Therefore the perturbation of CoM acceleration is used to simulate external disturbances.
This section consists of two parts.
In \cref{subsec:push_recovery} we present the simulation results when the robot faces an external push.
In \cref{subsec:comparison} we compare the two approaches.

\subsection{Push recovery of online approaches} \label{subsec:push_recovery}

\begin{figure}[t] \label{compos_push1}
  \centering
  \includegraphics[width = 0.45\textwidth]{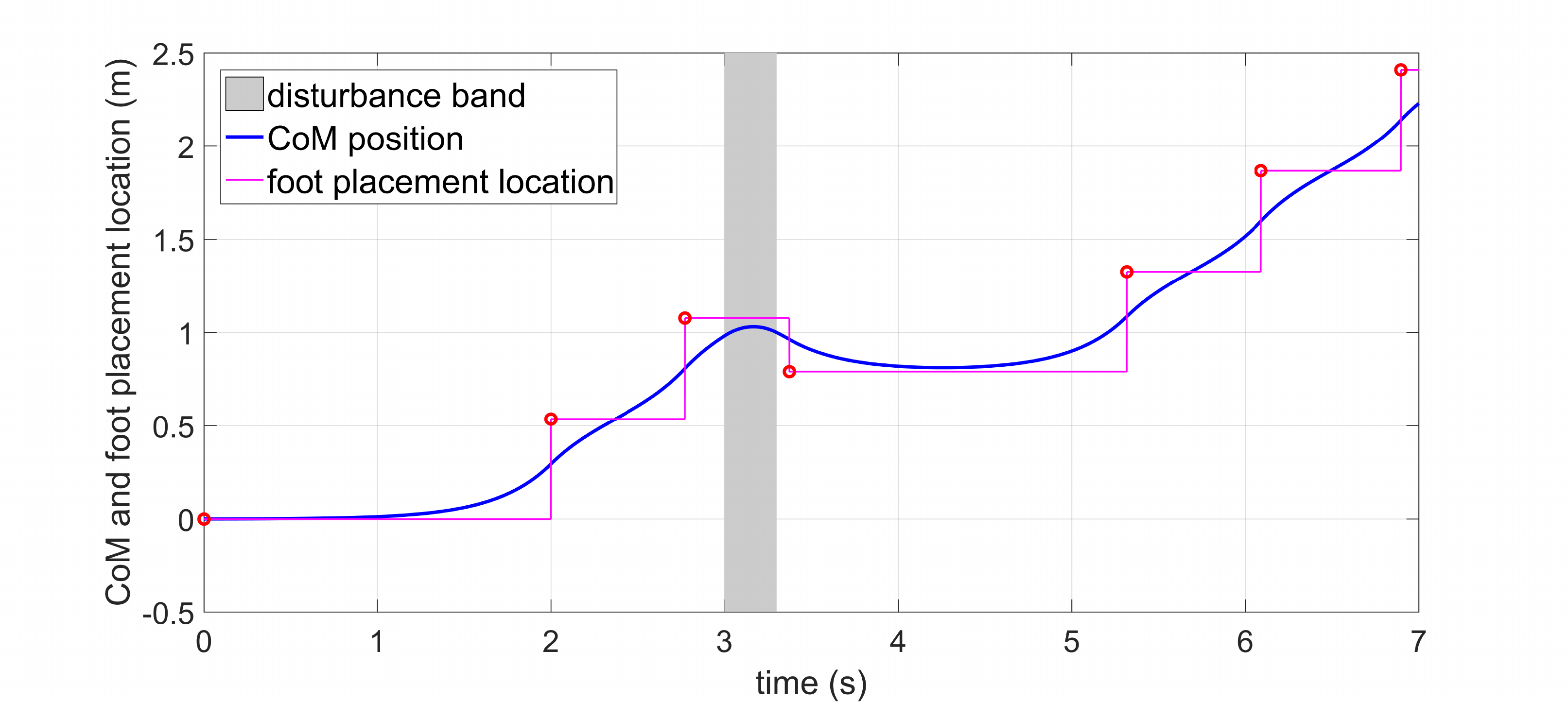}
  \caption{CoM and foot placement location in backward push recovery simulation.}
\end{figure}
\begin{figure}[t!] \label{T0_push1}
  \centering
  \includegraphics[width = 0.45\textwidth]{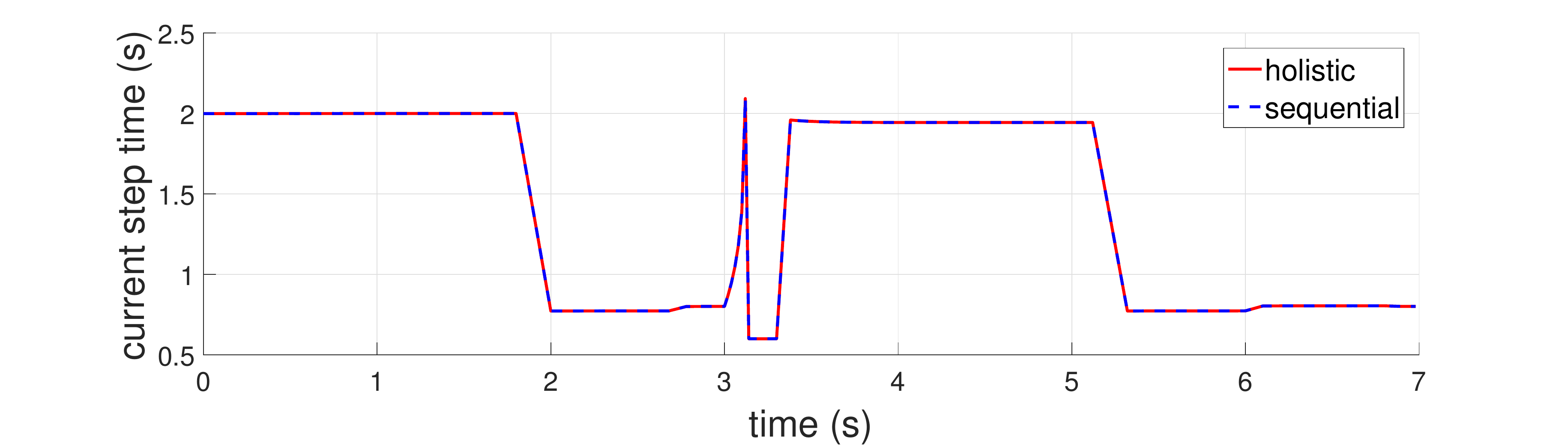}
  \caption{Current step duration in backward push recovery simulation.}
\end{figure}
\begin{figure}[t!] \label{T1_push1}
  \centering
  \includegraphics[width = 0.45\textwidth]{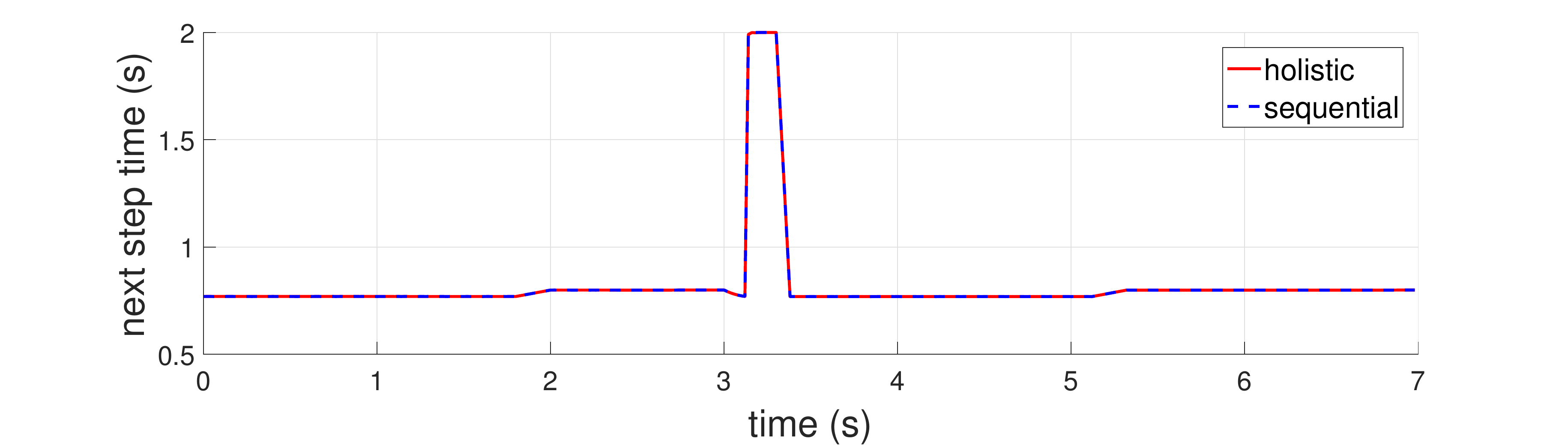}
  \caption{Next step duration in backward push recovery simulation.}
\end{figure}
\begin{figure}[t!] \label{p_push1}
  \centering
  \includegraphics[width = 0.45\textwidth]{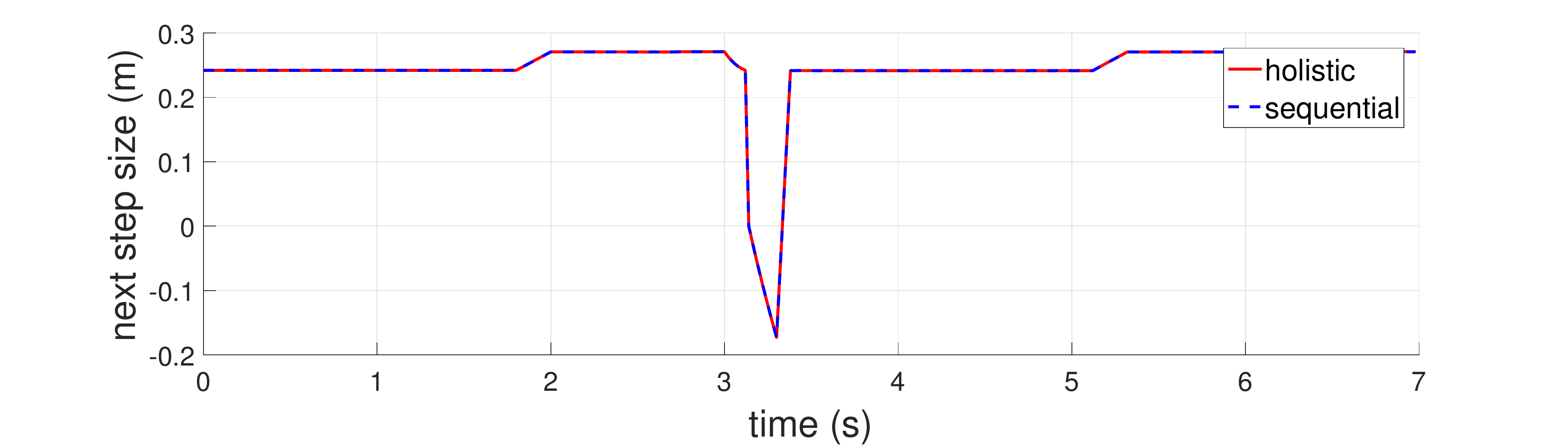}
  \caption{Next step size in backward push recovery simulation.}
\end{figure}

\begin{figure}[t] \label{compos_push2}
	\centering
	\includegraphics[width = 0.45\textwidth]{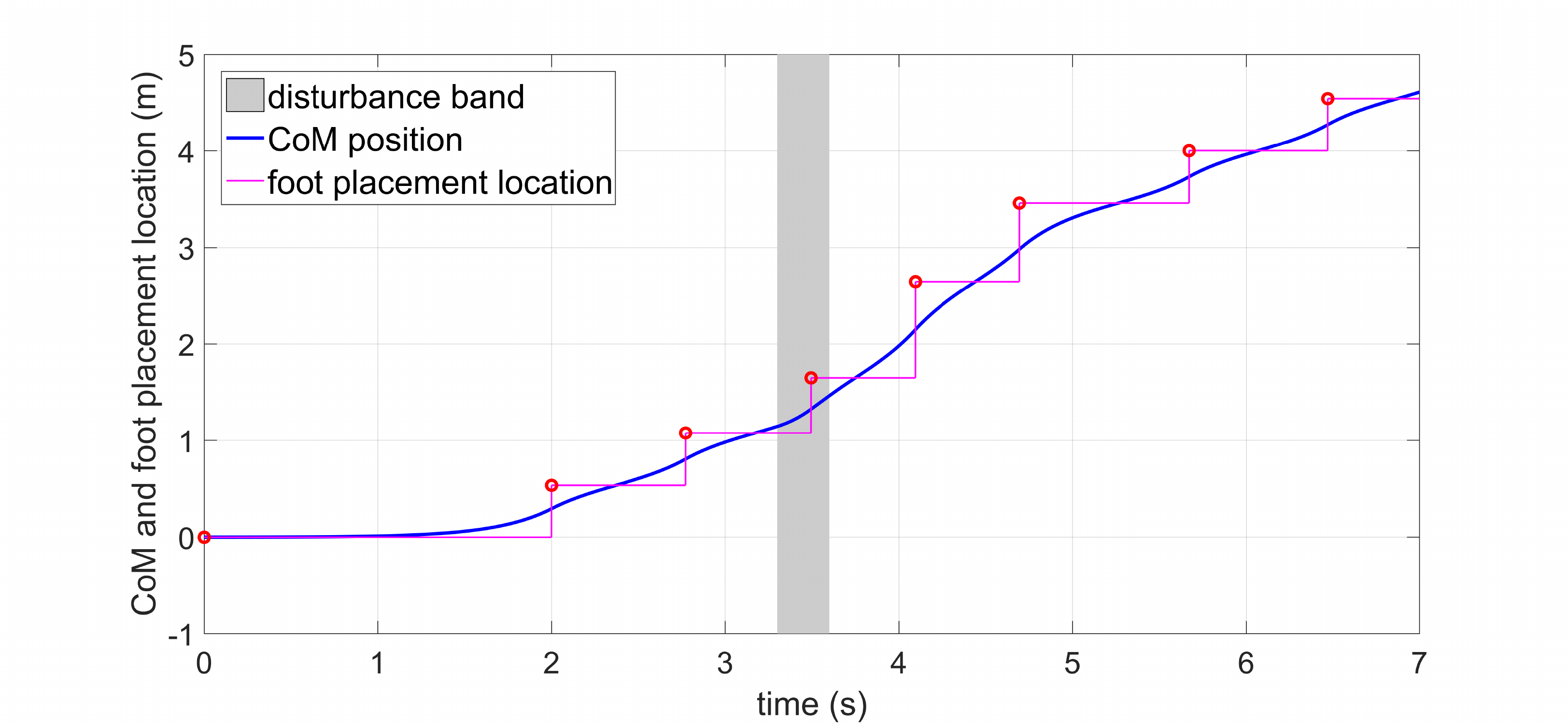}
	\caption{CoM and foot placement location in forward push recovery simulation.}
\end{figure}
\begin{figure}[t!] \label{fig:CoM_vel}
  \centering
  \includegraphics[width = 85mm]{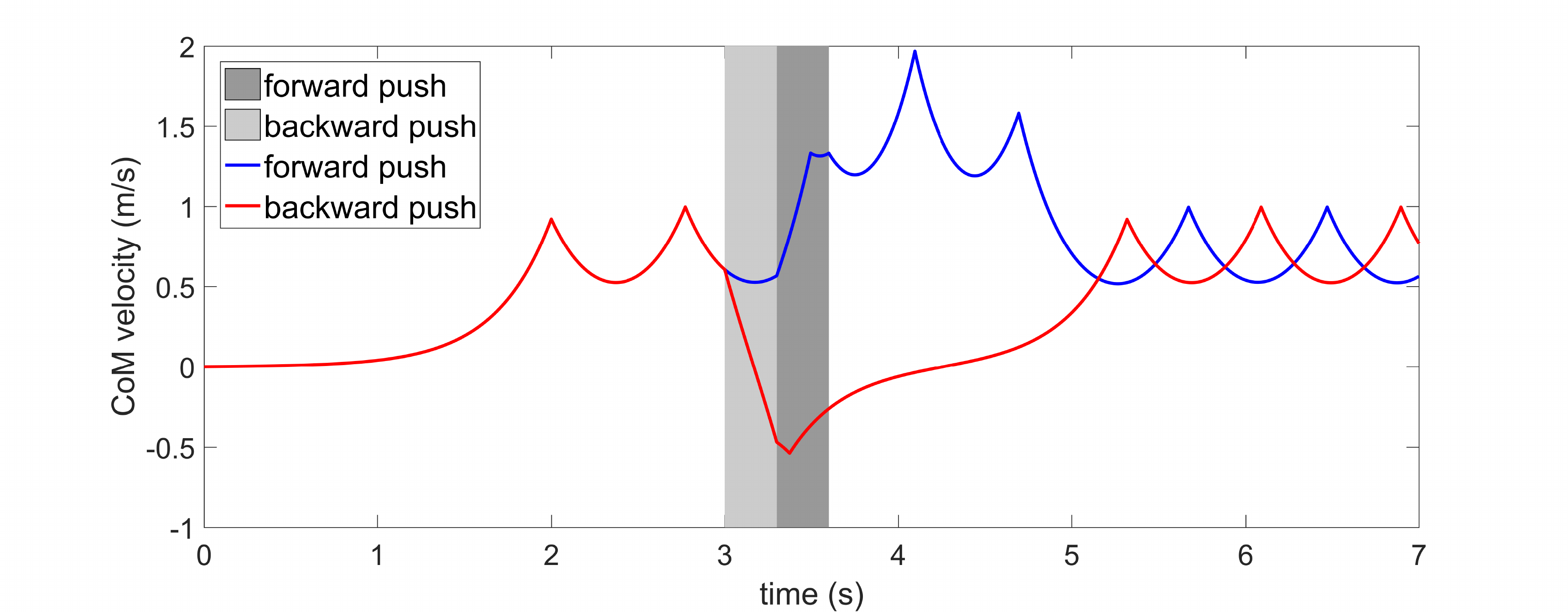}
  \caption{Velocity of CoM in both forward and backward push simulations. The velocity settles to the desired velocity roughly $2$s after the push.}
\end{figure}

In this section we present two typical push recovery settings -- a forward and a backward push -- as examples of how the robot maintains balance by adjusting the walking parameters.
For the simulations, the limits of the walking parameters were: $T_\text{min}=0.6s$, $T_\text{max}=2.0s$, and $L_\text{max}=0.5m$.
\cref{compos_push1} demonstrates the CoM trajectory and foot placement location during a backward push recovery and \cref{T0_push1,T1_push1,p_push1} demonstrate how the three walking parameters continuously change during the whole process.
The backward push occurs at $3.0s$ for $0.3s$ at a magnitude of $3.0m/s^2$.
From \cref{compos_push1}, it can be seen that when the robot encounters the push, the CoM slows down and reverses moving direction.
To prevent falling back, the robot ends the current step early and takes a small step back (\cref{T0_push1,T1_push1,p_push1}).
Then, after a long recovery period, the walking pattern resumes as planned.
In these three figures, the red solid line represents the parameters from the holistic approach, while the blue dashed line represents the parameters from the sequential approach. In this simulation, both approaches yield the same results.
\cref{compos_push2} demonstrates the forward push scenario.
The push happens at $3.3s$, lasting for $0.3s$ at a magnitude of $2.5m/s^2$.
In the following step, the robot reduces the step duration while increasing the step distance, and the CoM returns to the nominal state after $2s$. 
In these two simulations, the holistic and sequential approach result in the same walking parameters. \cref{fig:CoM_vel} demonstrates the fluctuation of CoM velocity under external disturbances.

\subsection{Comparison between two approaches}
\label{subsec:comparison}

\begin{figure}[t] 
  \centering
  \subfigure[Computation time of the holistic approach.] {
  \label{computation time holistic}
  \includegraphics[width = 0.45\textwidth]{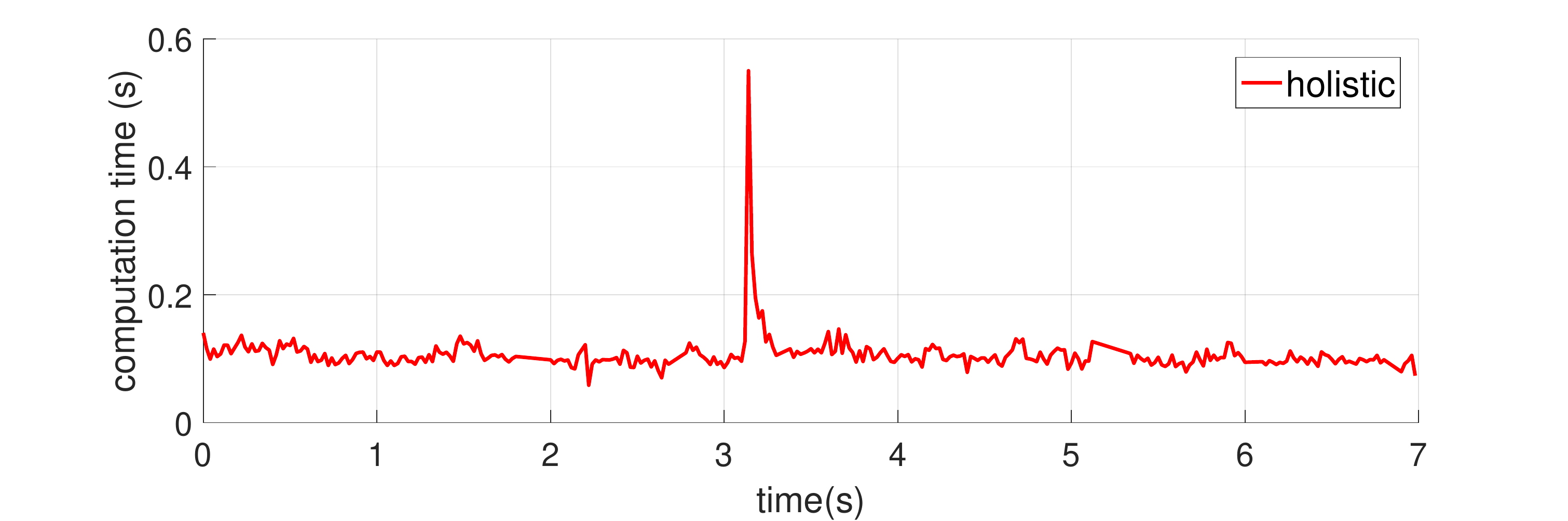}
  }
  \subfigure[Computation time of the sequential approach.]{
  \label{computation time sequential}
  \centering
  \includegraphics[width = 0.45\textwidth]{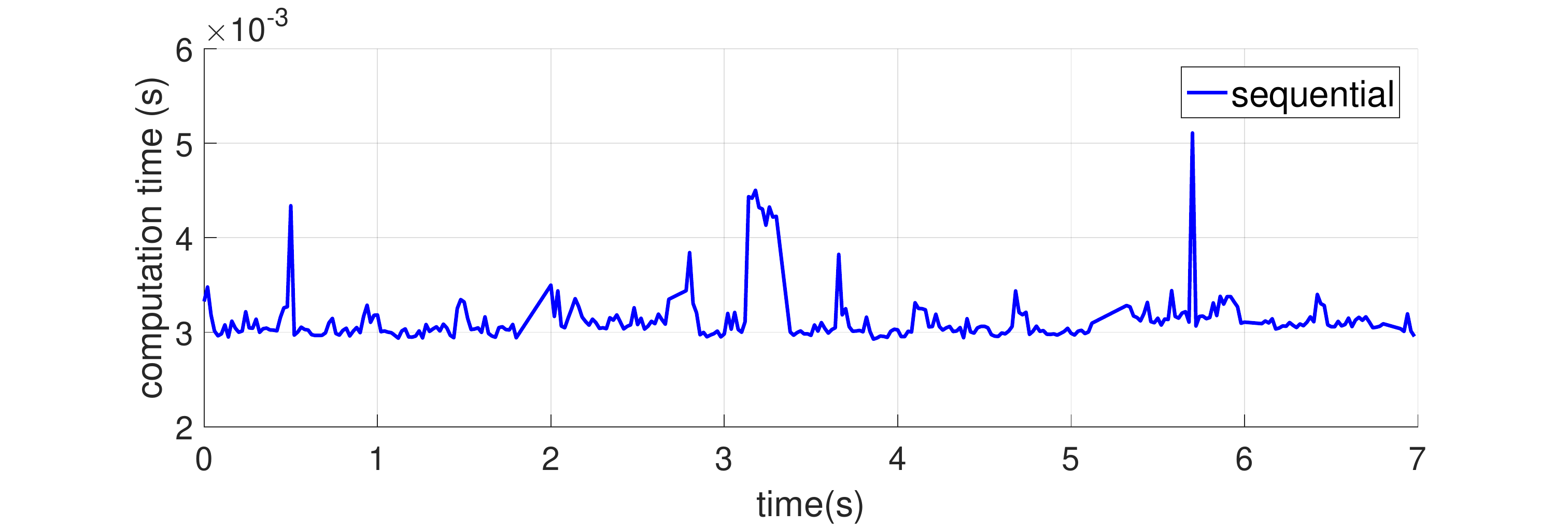}
  }
  \caption{Computation time of both methods during a backward push simulation discussed in \cref{subsec:push_recovery}.}
\end{figure}



\begin{figure}[t]
\label{ThreeDT0}
  \centering
  \subfigure[3-D figure of the optimal current remaining step duration at different discrete CoM states.] {
  \includegraphics[width = 0.45\textwidth]{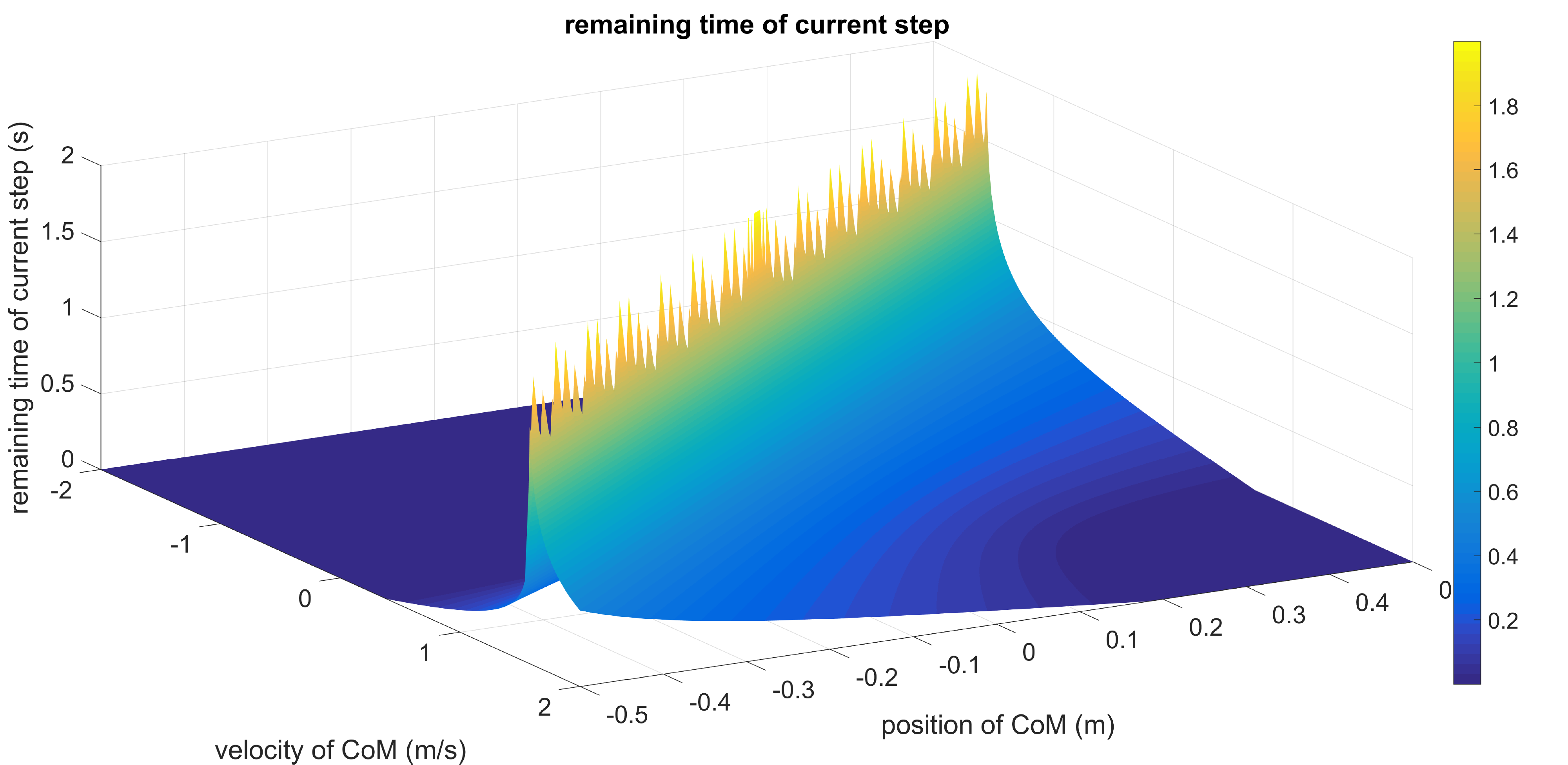}
  }
  
  \subfigure[3-D figure of the optimal next step duration at different discrete CoM states.]{
  \includegraphics[width = 0.45\textwidth]{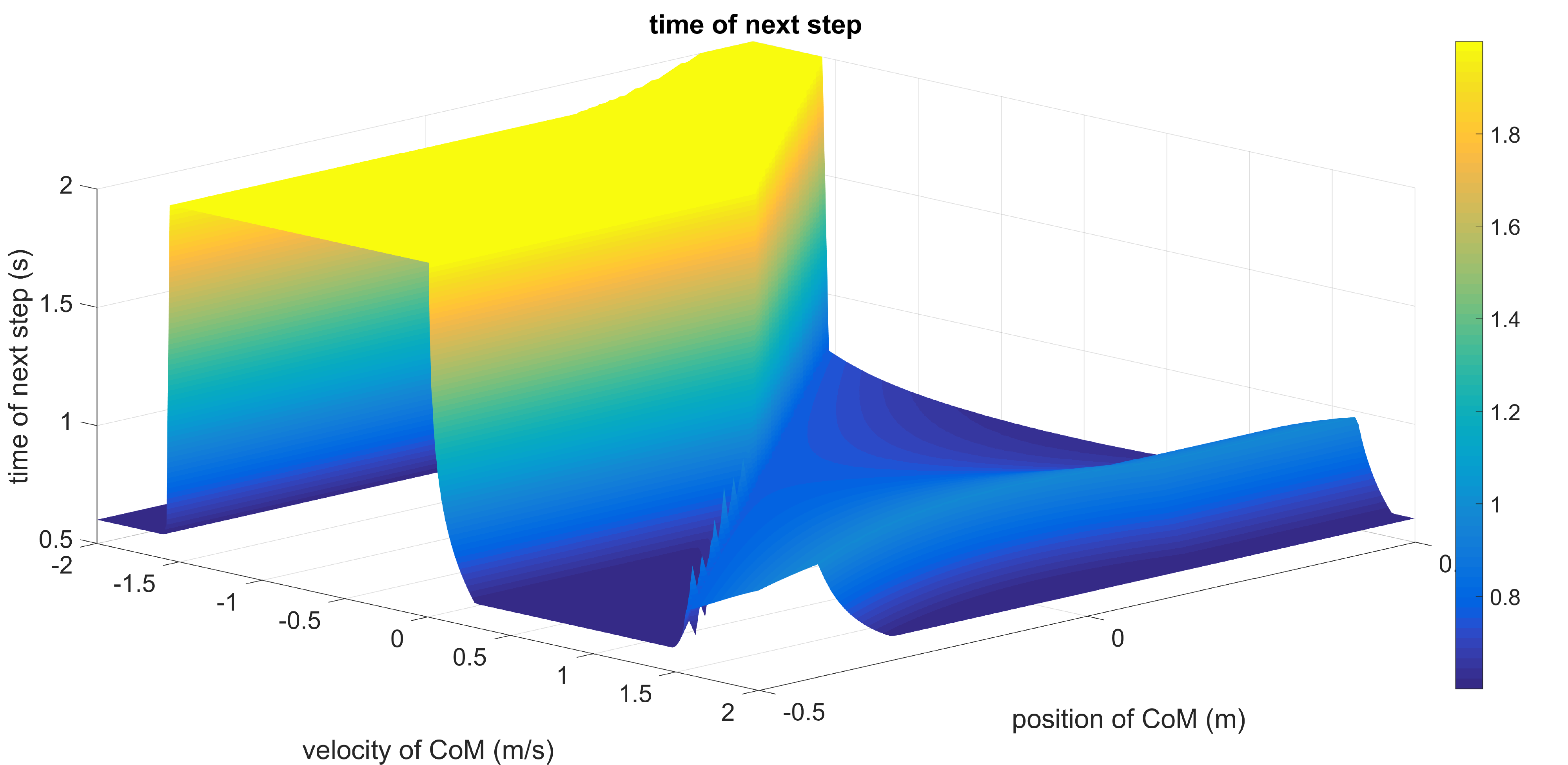}
  }
  \subfigure[3-D figure of the optimal next foot placement at different discrete CoM states.]{
  \includegraphics[width = 0.45\textwidth]{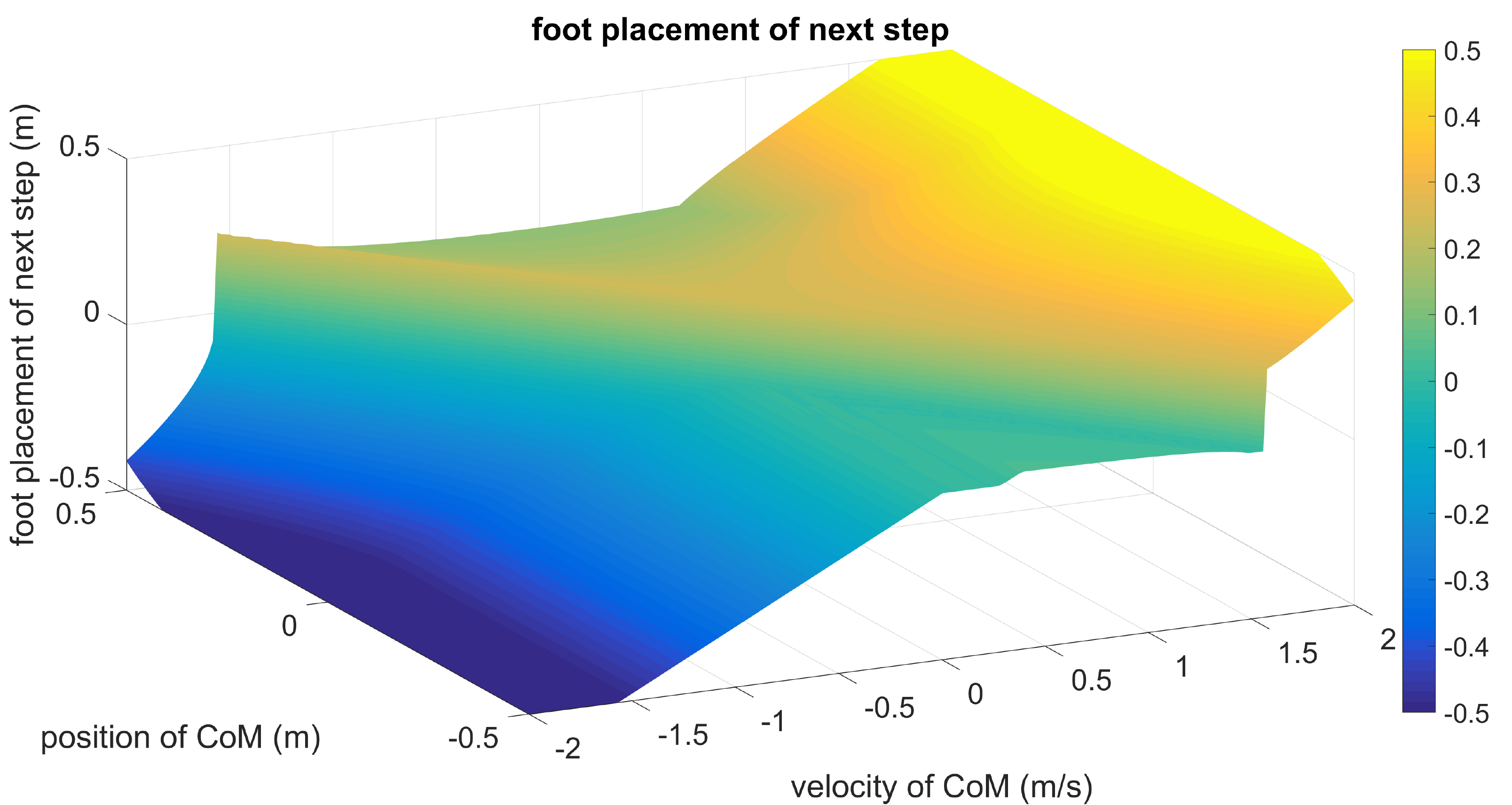}
  }
  \caption{Optimal control solutions at different CoM states from parameter scan of the CoM states using the holistic approach as a baseline.}
\end{figure}

\begin{figure}[t]
\label{seq-uni}
  \centering
  \subfigure[Costs from holistic approach for different discrete CoM states.] {
  \includegraphics[width = 0.45\textwidth]{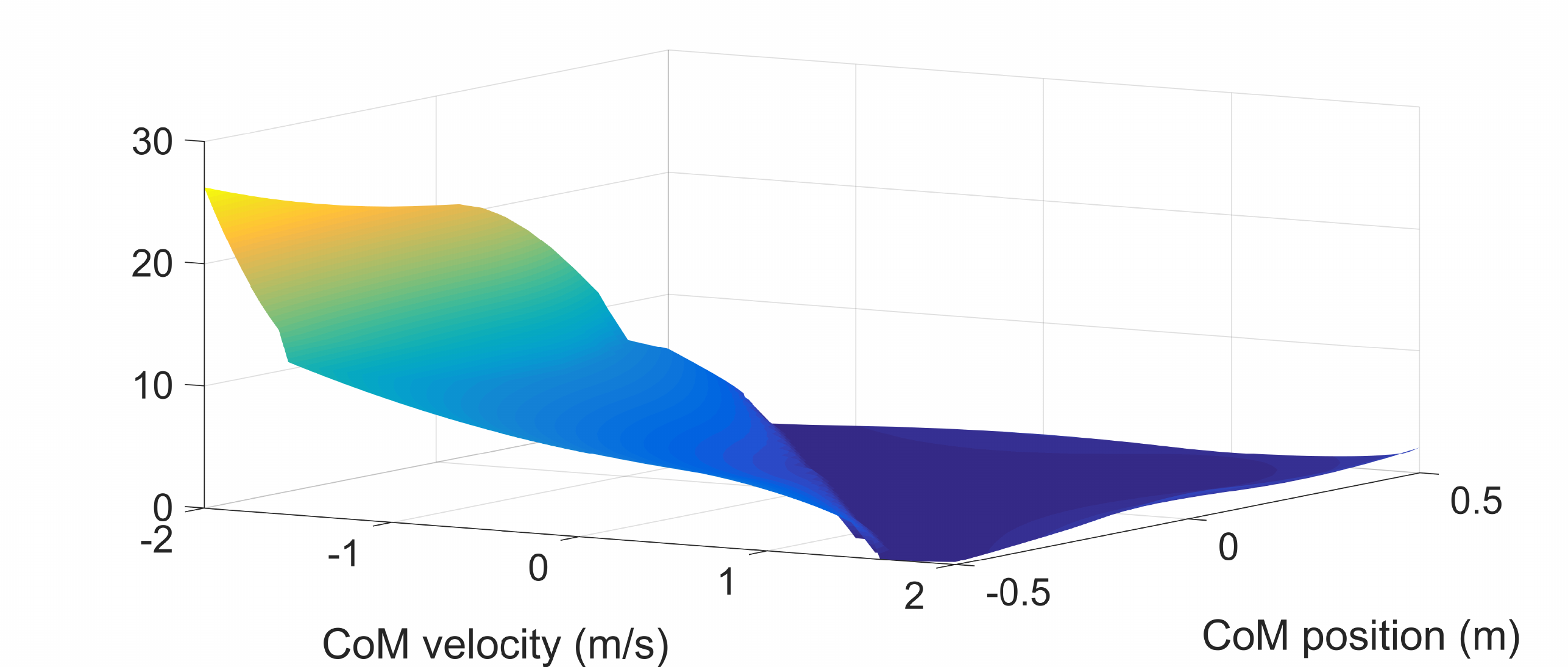}
  }
  
  \subfigure[Costs from sequential approach for different discrete CoM states.]{
  \includegraphics[width = 0.45\textwidth]{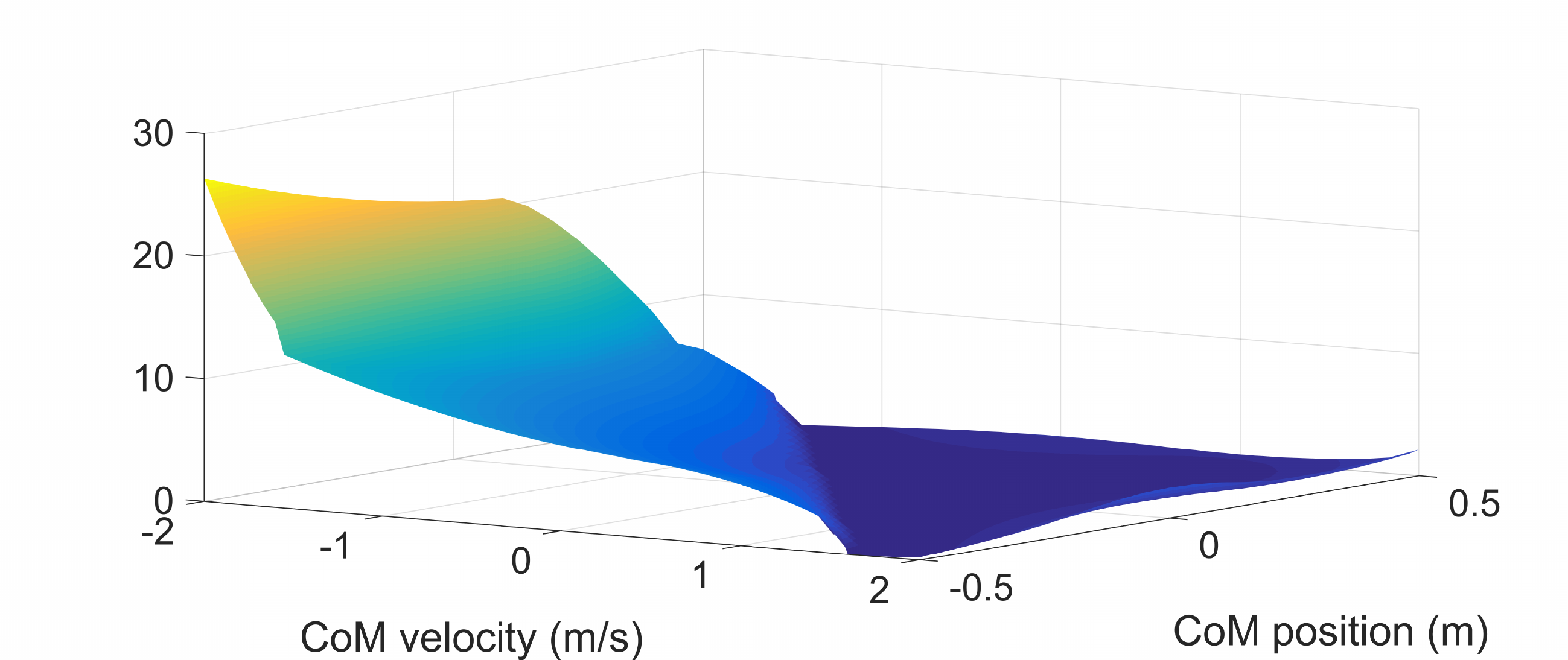}
  }
  \subfigure[Difference of costs: the sequential costs minus the holistic costs.]{
  \includegraphics[width = 0.45\textwidth]{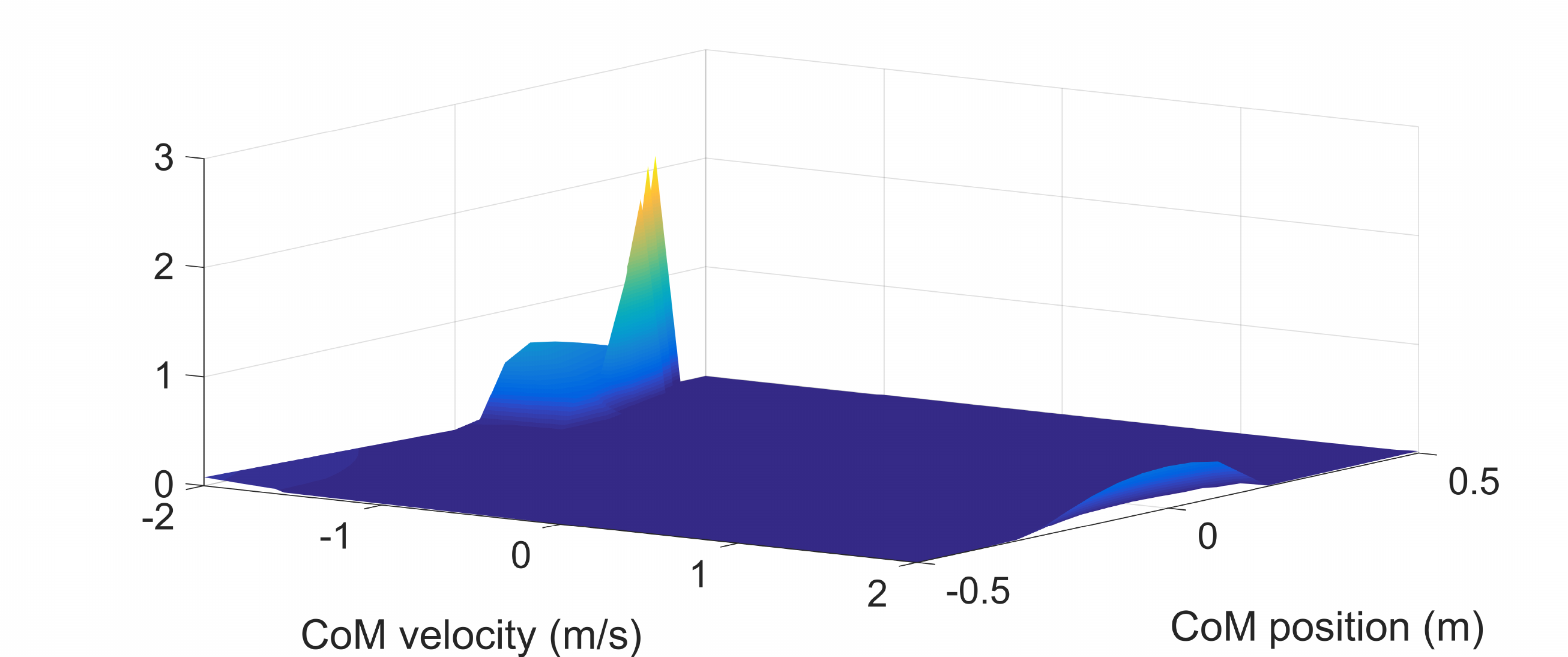}
  }
  \caption{Difference of costs between the two approaches.}
\end{figure}

%

\begin{table}[t!]
	\centering
	
  \caption{Example of current CoM state $\boldsymbol x_0 = [0.35, -1.90]^T$, where the two approaches behave differently for desired CoM state $\boldsymbol x_d = [0.27, 1.00]^T$.}
  \label{table: different between 2 methods}
  	\begin{tabular}{c|ccccc}
          Approach                & $T_0$ & $T_1$ &  $p^*$  &$\boldsymbol x_f^T$ & $\boldsymbol x_{\text{next}}^T$  \\
    \hline
    Holistic      & 0.17 & 2.00 & -0.50 &$[0.06, -1.56]$&$[0.29, 0.91 ]$			\\
    Sequential  \ & 0.08 & 0.60 & -0.50 &$[0.21, -1.69]$&$[-0.05, -0.66]$
\end{tabular}
\end{table}

Since the sequential approach reduces the number of optimization variables and divides the holistic three-parameter nonlinear optimization problem into separate one-parameter optimization problems, it calculates the optimal results much faster.
\cref{computation time holistic,computation time sequential} demonstrate the computation time of the two approaches in the forward push recovery simulation discussed in \cref{subsec:push_recovery}.
The average computation time of the holistic approach in this simulation is $112.6 ms$, while the computation time of the sequential approach is $3.4ms$.
On average, the holistic approach is approximately 30 times slower than the sequential.
The simulations are performed on a four-cored Intel Core i7 @ 2.6GHz computer using the standard MATLAB non-linear optimization solver \texttt{fmincon} (interior point algorithm).

In most situations, the two approaches are equivalent, because the error of the current step influences the next step motion.
Therefore, if the error of the final CoM state of the current step is minimized, then certainly a smaller deviation of next step duration and location is needed to restore to the desired state at the end of the next step.
Apart from the simulation performed in the previous section, we simulate several push recovery settings to analyze the performance of both approaches.
For different values of the time of impact, impact duration, magnitude, and direction of the external push, both approaches yield the same optimal parameters.

Even though the two approaches have similar behaviour, they exhibit some structural differences.
The sequential approach is a type of ``\textit{greedy}'' algorithm.
It only focuses on a single step in every stage, while the holistic approach takes 2 steps into consideration.
To understand better in which situations these two approaches will behave differently, we performed a \textit{global search} by discretizing the entire CoM states and computed the optimal walking parameters from two approaches.


We performed the parameter scan of the CoM state space and obtained the costs by the objective of \cref{eqn:holisitic_opt} for all discrete current CoM states. \cref{ThreeDT0} shows the 3D view of the optimal solution of each parameter in the COM state space obtained by the holistic approach. These are used as a baseline for the sequential approach to compare against. 

\cref{seq-uni} demonstrates the difference in the cost distribution between the sequential method and the holistic one. From \cref{seq-uni} we can infer that the two approaches obtain the same results for most CoM states, except some states located near the left and right edges of this figure, where the large negative CoM velocity happens near the friction cone, which is in fact an uncommon case. 

\cref{table: different between 2 methods} gives a representative example of different optimal results from the holistic and sequential approaches, where $T_0$, $T_1$, $p^*$ represent current remaining step duration, next step duration and next foot placement location respectively. The final state of the current step and the next step are $\bm{x}_f$, $\bm{x}_{\text{next}}$, respectively.
Since the sequential approach will minimize the error of the current final step state, in the sequential approach the robot will immediately put a step back in order to satisfy the desired CoM state, at the cost of a greater error in the next step.
In contrast, the holistic approach allows slightly larger errors in the current step while gaining a better minimization of the CoM state error at the end of the next step.


Even though the sequential approach does not necessarily recover a stable gait with a minimum error at the end of the next step, while the holistic approach can, it is difficult to conclude which approach is better for implementation.
The two approaches have slightly different strategies for the same CoM states.

Interestingly, the holistic approach is more likely to produce fewer long-duration steps to adjust, while the sequential approach generates more short-duration steps. In terms of actuation demand, the holistic approach has minimum efforts because of the less rapid reactions. But in turn, it has higher risks of falling because longer step duration puts the robot under the risk of modeling error and future disturbances. Meanwhile, this better optimality is traded off by more computational time. On the other hand, though the sequential approach does not render the global optimal all the time, its quick steps can be more advantageous in cases of multiple pushes and the little computational time is more suited for real-time control.

\section{Conclusion} \label{seq:conclusion}
In this paper, we present two online push recovery strategies utilizing nonlinear optimization, namely the holistic approach and the sequential approach, which are both capable of maintaining stability of bipedal walking based on the LIP model by adjusting the current remaining step duration, next step duration, and next step location.
Both approaches obtain mostly the same optimal results but differ to some extent at some CoM states.
Since the sequential approach splits the nonlinear problem of optimizing three parameters into a sequence of iterative one-dimensional searches, its computation speed increases significantly, making it 30 times faster than the holistic approach. Hence, it can be applied on real-time control systems, when computation power or time is limited.
Furthermore, the robustness of our approaches were demonstrated by applying various pushes, and the similarities and small differences are compared.

Our future work will extend these approaches to three-dimensional scenarios by including the three-dimensional constraints of the foothold location as well as the friction cone.
The over-simplicity of the robot dynamics introduced by the LIP model can be a hurdle in a real robot, so another extension of our work is to utilize a more complex multi-link robot model similar to the work in \cite{you2016foot}.

\section*{Acknowledgement}
This work is supported by Future AI and Robotics Hub for Space (EP/R026092/1) and UK Robotics and Artificial Intelligence Hub for Offshore Energy Asset Integrity Management (EP/R026173/1) funded by the EPSRC. 

\addtolength{\textheight}{-12cm}   


%



\end{document}